\documentclass[twocolumn,10pt]{IEEEtran}

\usepackage{color}

\usepackage{rotating}

\usepackage[colorinlistoftodos,prependcaption,textsize=tiny]{todonotes}

\usepackage{cite}      

\usepackage{graphicx}  

\usepackage{textcomp}

\usepackage{amsmath}   

\usepackage{amssymb}

\usepackage{float}

\usepackage{ifpdf}

\usepackage[]{algorithm2e}

\usepackage{tabularx,colortbl}
\usepackage{xcolor}

\definecolor{darkred}{rgb}{0.55, 0.0, 0.0}
\definecolor{darkblue}{rgb}{0.0, 0.45, 0.95}
\definecolor{darkgreen}{rgb}{0.0, 0.55, 0.0}

\usepackage{titlesec}

\begin{document}
%
\title{Spatial Normalization for Cross-Domain Retinal Layer Segmentation in Optical Coherence Tomography}

%
\author{Iker Moran-Cavero, Monica Hernandez, Elvira Mayordomo, Naiara Artiaga, Beatriz Pardiñas, Beatriz Cordon and Elena Garcia-Martin 
}
%
%

\maketitle

\markboth{Draft submitted to Journal of Biomedical and Health Informatics (JBHI-****-****)}{Moran \MakeLowercase{\textit{et al.}}: ANTs OCT Seg}
%



\begin{abstract}

Retinal layer segmentation in Optical Coherence Tomography (OCT) is a fundamental step for extracting
quantitative biomarkers of retinal structure. Indeed, there is a growing interest in the analysis of OCTs in the context
of neurodegenerative diseases. However, segmentation remains challenging due to speckle noise, shadowing artifacts, low contrast between adjacent
layers, anatomical variability across subjects, and domain shifts arising from different acquisition protocols
and clinical populations. While deep learning methods have achieved remarkable performance, their robustness
and generalization across heterogeneous datasets remain limited.
In this work, we investigate the role of spatial normalization as a preprocessing strategy to mitigate geometric domain shifts
and improve the consistency of retinal layer segmentation. Inspired by standard practices in neuroimaging, we introduce a fovea-centered normalization
framework that aligns OCT volumes into a common anatomical reference. We perform a comprehensive evaluation of state-of-the-art deep learning architectures.
To provide a comprehensive assessment of segmentation quality, we combine conventional overlap-based metrics at B-scan level
with topology-aware metrics at A-scan level and thickness-based measures at the en-face level.
In cases where a ground truth is not available, we propose topology violation quantitative metrics that do not require ground truth annotations
and a thickness-based qualitative assessment that captures structural consistency and clinically relevant patterns at the en-face level.
The results demonstrate the importance of spatial normalization in OCT segmentation pipelines toward the development of robust and clinically meaningful
retinal analysis tools, enabling reliable biomarker extraction and downstream computational analysis in neurodegenerative research.
\end{abstract}


%
\IEEEpeerreviewmaketitle


\section{Introduction}

\subsection{Context}

Optical Coherence Tomography (OCT) has become an essential imaging modality in clinical ophthalmology, providing a fast and non-invasive
visualization of the retinal structure. Its widespread adoption in routine practice has enabled detailed quantitative analysis of retinal
layers, supporting the assessment of both normal retinal anatomy and pathological alterations. Retinal layer segmentation
constitutes a key step for extracting clinically meaningful structural measurements from OCT images~\cite{Viedma_22,Zhang_25}.

Beyond the established role of OCT in ocular diseases, increasing interest has been directed toward the potential of retinal imaging as an
indirect marker of neurological and neurodegenerative processes. Several studies have reported associations between alterations in
specific retinal layers and neurodegenerative conditions~\cite{Knier_15,Lapiscina_16,London_19,Orbus_22,Hernandez_23}.
However, the extent to which retinal measurements can reliably reflect
central nervous system pathology remains an open question. Accurate and robust segmentation of all retinal layers is therefore a prerequisite
for further investigating these relationships across different subject populations.


Retinal layer segmentation in OCT images remains a challenging task due to a combination of factors that can be broadly classified into
imaging-related issues and anatomical sources of variability~\cite{Zhang_25}.
From an imaging perspective, OCT data are inherently affected by speckle noise, which reduces contrast and blurs layer boundaries.
In addition, shadowing artifacts caused by retinal vessels or highly reflective structures can locally attenuate or distort the signal,
which may result in missing, poorly defined, or strongly deformed layer interfaces.
Several retinal layers are closely spaced and exhibit subtle intensity differences, making their separation difficult,
particularly when the axial or lateral resolution is insufficient to clearly resolve thin structures in certain slices.

From an anatomical perspective, substantial inter-subject variability can be observed in retinal thickness, curvature, and reflectivity profiles.
This variability appears in both healthy and diseased populations, and it may be further amplified in pathological conditions, posing a significant
challenge for automated segmentation methods~\cite{Ran_21,Lee_23}.
Deep learning-based segmentation methods are particularly sensitive to such anatomical variations as well as to the known shifts in data distribution caused
by differences in acquisition protocols, imaging devices, or preprocessing strategies.
As a result, model performance can degrade when confronted with unseen anatomical configurations or domain changes.
Although spatial data augmentation can partially compensate for this variability, it does not explicitly resolve
anatomical misalignment across subjects, and generalization to unseen domains remains limited.

Spatial image normalization is a fundamental stage in many medical imaging applications.
This process is used to align images to a common anatomical reference, reducing inter-subject variability.
Spatial normalization is critical in neuroimaging studies, enhancing the consistency of segmentation results, facilitating statistical analyses, and allowing direct comparison of structural metrics across
populations~\cite{Hadj_16,Tustison_21}.
Beyond standard intensity normalization, spatial normalization may be employed to align OCT images to a common anatomical reference, mitigating spatial domain shifts and
offering additional benefits for deep-learning retinal layer segmentation such as stabilizing network training, improving convergence, and enhancing generalization.


\subsection{Related Work}

Retinal layer segmentation in OCT has been addressed using a wide range of methods, each tailored to cope with the distinct structural and pathological
characteristics present in clinical data~\cite{Zhang_25}.
Many existing works implicitly assume a relatively regular retinal anatomy~\cite{He_19,He_19b,Li_21,Fazekas_22} or focus on specific pathologies~\cite{Bogunovic_19,Xing_22,Fazekas_26}.
In this study, we exclude complex retinal conditions such as
drusen, advanced macular degeneration, intraretinal fluid, macular edema, epiretinal membranes, glaucoma,
and other structural alterations that require specialized handling beyond layer delineation.
The segmentation challenges described in the previous section have motivated both classic and deep learning-based strategies that aim to delineate layer
boundaries with high accuracy and robustness.

Traditional graph‑based segmentation algorithms were among the first automated techniques developed for OCT layer delineation~\cite{Garvin_09,Chiu_15}.
These methods typically formulate the segmentation problem as an optimization over a weighted graph, where layer boundaries correspond to minimum‑cost paths.
Graph search and shortest path approaches have been shown to segment multiple intraretinal layers with accuracy comparable to expert graders,
and similar algorithms have been integrated into commercial OCT devices to provide device‑independent thickness measurements and fast layer delineation for
routine analysis~\cite{Garvin_09}.
Despite their interpretability and reproducibility, such optimization‑based methods rely on handcrafted features and cost functions,
and consequently exhibit limited generalization when applied to scans with substantial intensity differences or anatomical deviations from normative populations.

With the emergence of deep learning, neural network–based models have become predominant in retinal layer segmentation from OCT.
Fully convolutional networks, UNet architectures, and regression networks have been applied to learn pixel‑wise layer labels or surface positions directly
from training data, often yielding superior performance over classical methods~\cite{He_21}.
Among the most widely studied architectures, we should mention HeNet, MGU‑Net, and SD‑LayerNet.
HeNet models the problem with surface positions and employs hierarchical feature extraction to capture both
local and global contextual information, improving the delineation of closely spaced layers~\cite{He_19, He_21}.
MGU‑Net integrates multi-scale guidance in a graph convolutional network and residual connections
to refine boundary predictions, enhancing robustness against imaging noise and anatomical variability~\cite{Li_21}.
SD‑LayerNet introduces semi-supervised learning combined with anatomical priors and disentangled representations,
enabling accurate segmentation while preserving topological consistency~\cite{Chartsias_19,Fazekas_22}.
These architectures have demonstrated significant improvements over traditional methods in terms of accuracy,
topological correctness, and generalization to diverse OCT datasets.

In addition, several specializations of the SD-LayerNet architecture have been proposed to address specific
limitations of the baseline model, such as SD-LayerNet2 with a differentiable topological engine enabling
the simultaneous use of 1D and 2D representations of the retinal layers and anatomical priors intended to
improve robustness to scans with significant tilt~\cite{Fazekas_25}, or
SD-RetinaNet, specialized in joint layer and lesion segmentation~\cite{Fazekas_26}.


In addition to full layer delineation models, simpler segmentation strategies that group several consecutive layers into a single label have also been explored~\cite{Ma_21,Gende_23,Alvarez_24}.
Specifically, the studies in~\cite{Gende_23,Alvarez_24} use models derived from MGU-Net~\cite{Li_21} and nnU-Net~\cite{Isensee_21} and focus on isolating the retinal nerve fiber layer (RNFL) and grouping from the Ganglion Cell
Layer (GCL) to the Retinal Pigment Epithelium (RPE) without resolving all individual layers in the macular area.
This idea may be of interest in animal OCT imaging, where imaging resolution is usually limited, and coherently grouping retinal layers can improve the reliability of thickness measurements while simplifying the segmentation task.
For instance, RetOCTNet was developed to segment RNFL and overall retinal thickness in animal models, demonstrating high consistency with manual annotations~\cite{Sanchez_25}.

However, in human studies focused on neurodegenerative conditions, interest is often centered on the ganglion cell layer
and the inner plexiform layer (GCL and IPL) of the macular area,
and an overly aggressive grouping of layers including GCL and IPL may obscure relevant structural changes.
While simplified aggregation can be effective for targeted or low-resolution applications, human OCT studies would benefit from more granular segmentation schemes that
preserve critical layers, enabling detailed quantitative analyses and facilitating the detection of subtle disease-related alterations.

The scarcity of fully annotated OCT datasets has been acknowledged as the major limitation in the development of retinal layer segmentation algorithms~\cite{Li_24,Zhang_25}.
Manual delineation of all intraretinal layers is labor-intensive and requires expert graders, resulting in limited availability of high-quality, multi-layer annotated volumes.
This limitation is particularly evident in the peripapillary region, where although datasets involve 2D annotations,
they cover small cohorts, and are rarely publicly available. In addition, these datasets are largely restricted to
glaucoma studies and are known to suffer from annotation inconsistencies or limited reliability.
Architectures such as HeNet and MGU‑Net, while effective at segmenting multiple layers, rely on fully supervised training, and they do not incorporate mechanisms to address the shortage of annotated data.
SD-LayerNet explicitly tackles this challenge by employing a semi-supervised framework, enabling accurate segmentations even with relatively few labeled volumes.
Despite these efforts, the lack of large, fully annotated OCT datasets remains a critical bottleneck for developing robust and generalizable retinal segmentation methods.

Across these classical and deep learning approaches for OCT segmentation, a notable gap in the literature is the lack of systematic investigation into spatial preprocessing strategies such as anatomical alignment or normalization~\cite{Hadj_16,Tustison_21}.
Although intensity normalization and data augmentation are commonly used to mitigate domain shift and enhance generalization, there are no comprehensive studies evaluating how bridging the gap to spatial normalization in OCT
might improve the performance of segmentation models and, ultimately, benefit downstream analysis tasks such as retinal layer thickness estimation, biomarker extraction, or morphometry studies.



\subsection{Contribution}

% ...

In this work, we present a comprehensive comparative study of state-of-the-art deep learning architectures for retinal layer segmentation in OCT, including nnU-Net~\cite{Isensee_21}, MGU-Net~\cite{Li_21}, and SD-LayerNet~\cite{Fazekas_22}.
Our evaluation focuses on the performance of these methods when trained on a relatively small dataset with limited ground truth that differs from the test distribution.
Specifically, the training data consists of OCT scans from Johns Hopkins University controls and multiple sclerosis patients~\cite{He_19}.
The test set includes scans of two different protocols of a clinically acquired dataset spanning over a decade at Miguel Servet University Hospital,
which includes healthy controls, patients with neurodegenerative diseases, and individuals with Steinert disease, reflecting a broad spectrum of retinal variability encountered in routine practice.
This evaluation setup introduces significant intensity and geometric domain shifts, challenging the generalization capabilities of the segmentation models.

We show that the choice of a preprocessing based on anatomical normalization for training and inference plays a critical role in achieving robust and anatomically consistent segmentations across subjects and scans.
To this end, we introduce a fovea-centered normalization strategy in which volumes are aligned by centering the fovea and rotated to place the retinal pigment epithelium (RPE) of the central scan along a horizontal reference axis.
Our results highlight that anatomical alignment combined with either supervised or semi-supervised versions of SD-LayerNet provides a superior framework for retinal layer segmentation, offering improved generalization capability, structural coherence,
and robustness without requiring large numbers of OCT volumes for training.
Notably, the labeled data used for training comprises only 35 volumes with approximately 50 B-scans per patient, yet our approach achieves high-quality anatomically consistent segmentations.

Indeed, we propose two complementary evaluation methods that enable a more comprehensive assessment of segmentation quality without relying on ground truth annotations:
topology preservation quantitative metrics computed at A-scan level,
and en-face integrity qualitative evaluation of the consistency of the different layer thicknesses.
The results in this work can guide the selection of segmentation strategies that ensure reliable, anatomically coherent outputs, enabling robust and reproducible retinal measurements for clinical
evaluation and quantitative analysis.

%

\section{Materials and Methods}
\label{sec:MatsMeths}


\subsection{Datasets}

\subsubsection{OCT acquisition protocols}

\begin{figure*}
\centering
\begin{tabular}{ccc}
\includegraphics[width=4 cm, height = 3 cm]{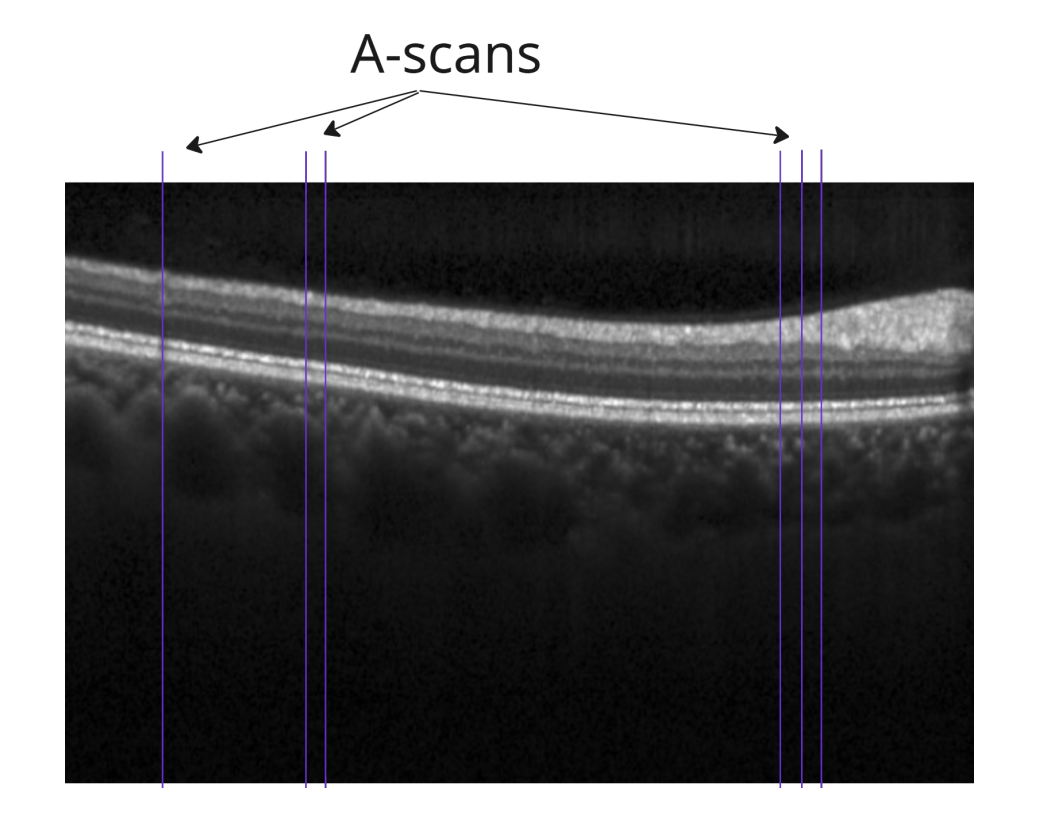} &
\includegraphics[width=3.5 cm, height = 3 cm]{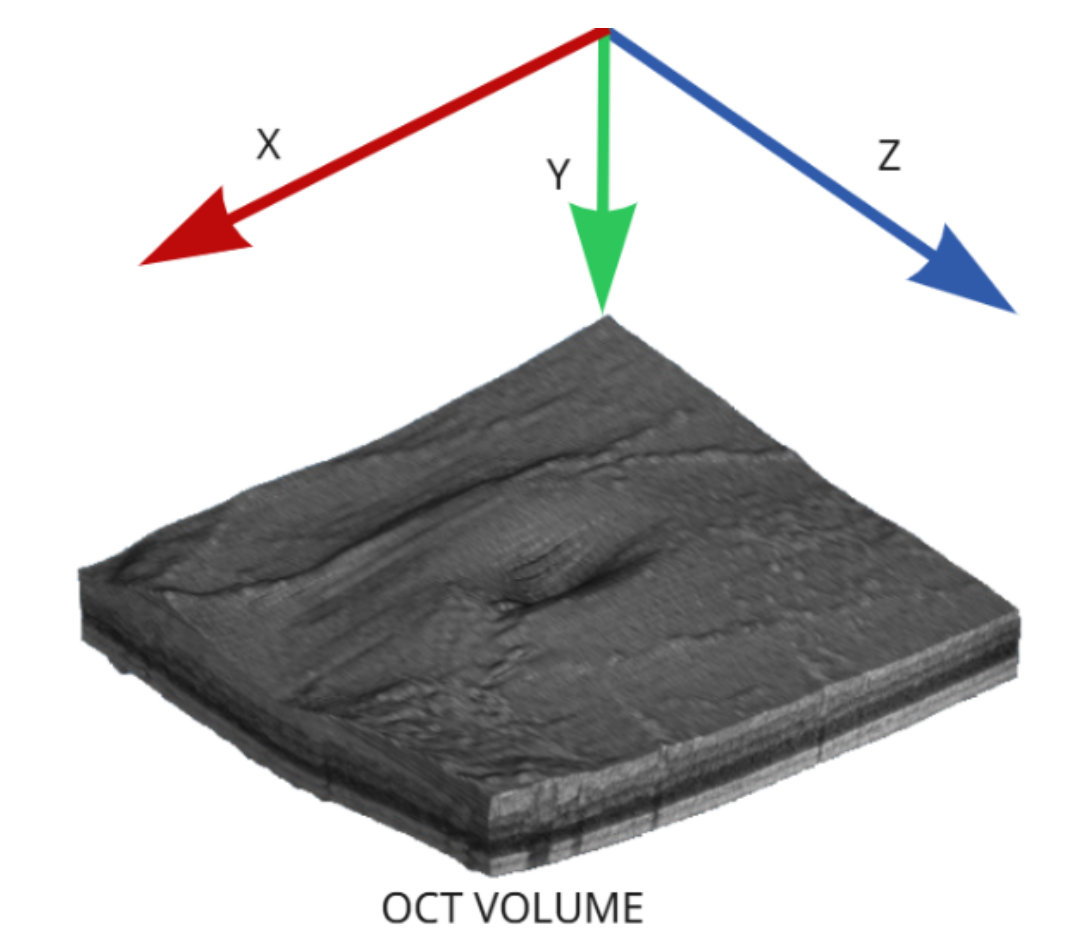} &
\includegraphics[width=4.5 cm, height = 3 cm]{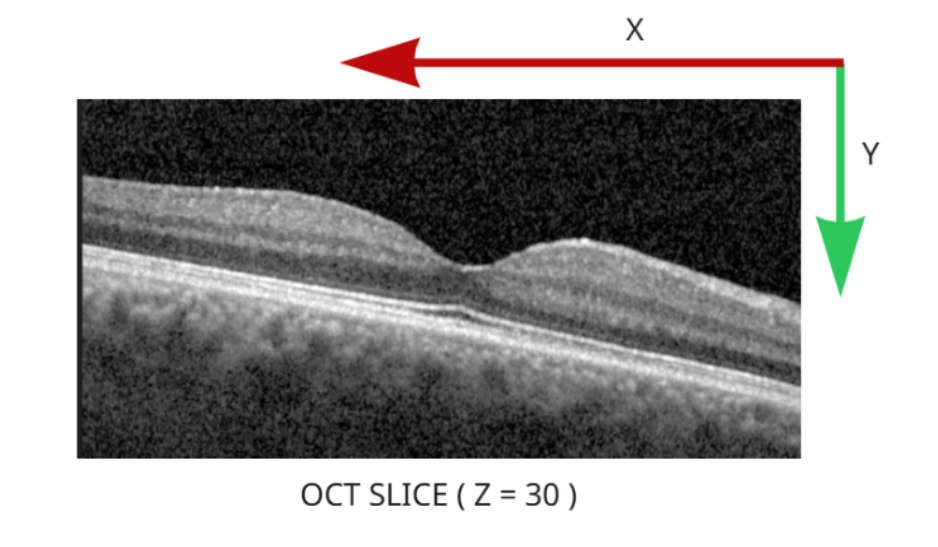} \\
\end{tabular}
 \caption{OCT acquisition units. Left, slice of a B-scan made of A-scans.
 Center, rendering of an OCT volume made of B-scans.
 Right, B-scan and X-Y coordinate system.
 }
 \label{fig:OCTScans}
\end{figure*}

In OCT imaging, the fundamental acquisition unit is the A-scan, a one-dimensional axial depth profile obtained from a single lateral position on the retina.
Each A-scan encodes the reflectivity of the tissue as a function of depth, revealing the layered retinal structure along that line of sight.
By laterally sampling many adjacent A-scans, the device constructs a B-scan,
a two-dimensional cross-sectional image that provides a slice of the retina with both axial and lateral information.
A volumetric OCT scan is formed by acquiring a series of B-scans across the region of interest.
En-face representations are derived by aggregating volumetric information along the axial direction, producing two-dimensional maps that
describe the spatial distribution of retinal structures across the imaging plane.
Figure~\ref{fig:OCTScans} shows examples of A-scan, B-scan, and volumetric OCT.

\begin{figure*}
\centering
\begin{tabular}{cc}
\includegraphics[height=2.75 cm]{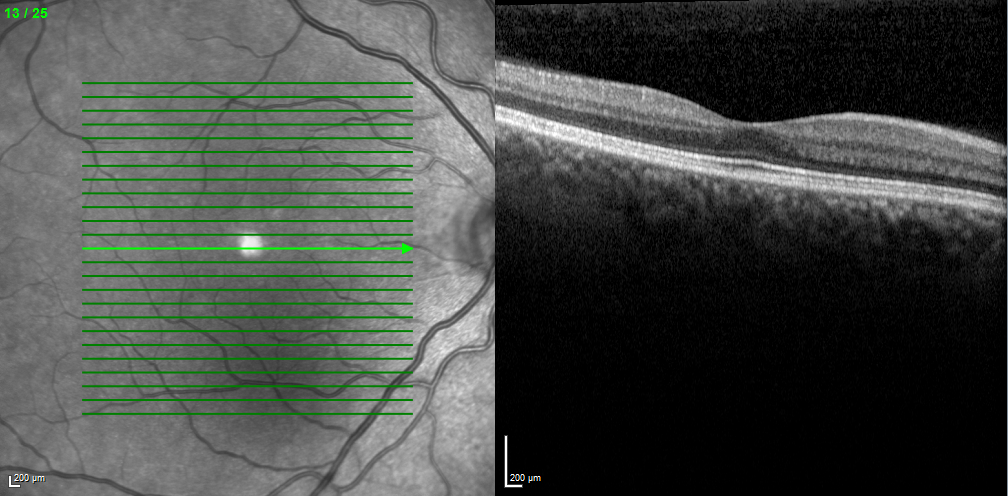} &
\includegraphics[height=2.75 cm]{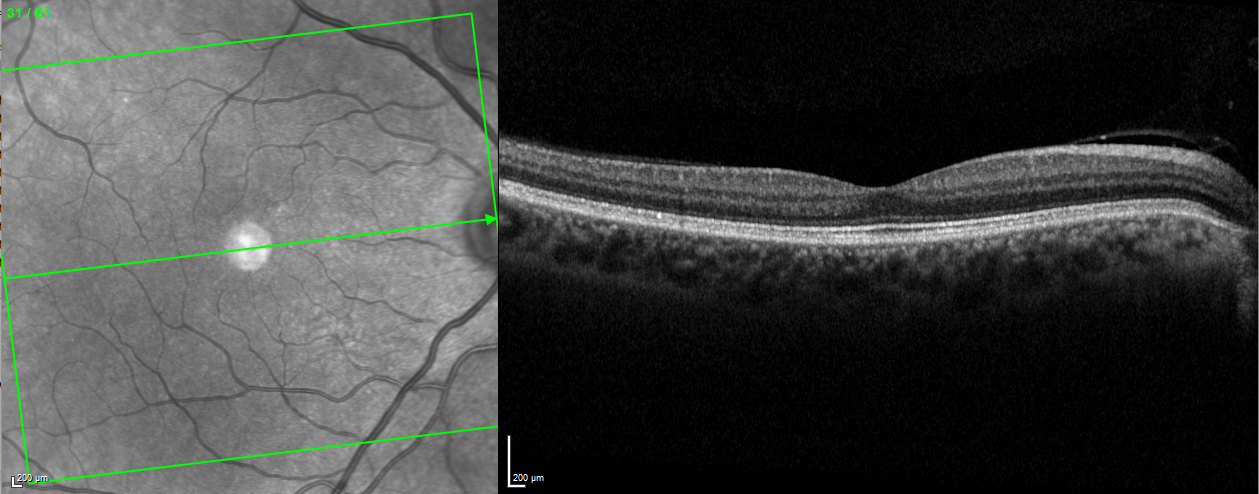} \\
\end{tabular}
 \caption{OCT protocols.
 Retinographies and corresponding fovea-crossing B-scans for the OCT protocols used in this study.
 Fast Macula (FastMac) protocol (left) and Posterior Pole (PPole) protocol (right).
 Images correspond to the same subject.
 For each pair, the arrow in the retinography on the left indicates the direction of the B-scan displayed on the right.
 Alignment between the fovea and the optic nerve head can be appreciated in PPole, while not in FastMac.
 }
 \label{fig:OCTProtocols}
\end{figure*}

In this work, we focus on two OCT acquisition protocols centered on the macular region.
These protocols are standard in the Spectralis OCT system from Heidelberg Engineering.
Figure~\ref{fig:OCTProtocols} shows an example of both protocols in the same subject.
The Fast Macula (FastMac) protocol consists of a rapidly acquired volumetric scan composed of a set of B-scans centered on the fovea,
providing sufficient spatial resolution for routine macular assessment while minimizing acquisition time.
The Posterior Pole (PPole) protocol captures a larger and systematically distributed rectangular grid of B-scans spanning the posterior pole,
enabling consistent thickness mapping of macular and peripapillary layers over a wider field of view. The PPole protocol benefits from the
Automatic Positioning System (APS), which standardizes scan placement by aligning acquisitions with anatomical landmarks such as the fovea and
the center of the Bruch’s membrane opening (BMO), which points out the anatomical center of the optic nerve head, thereby improving reproducibility
and reducing inter-session variability.

\subsubsection{Johns Hopkins dataset}

\begin{figure*}
\centering
\begin{tabular}{cccc}
\includegraphics[height=2.75 cm]{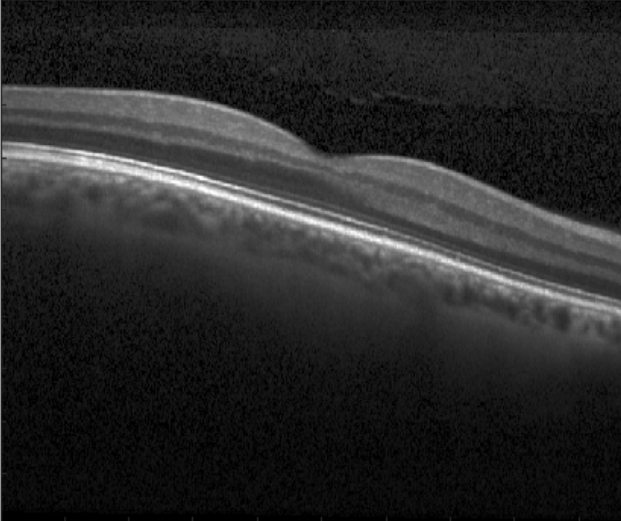} &
\includegraphics[height=2.75 cm]{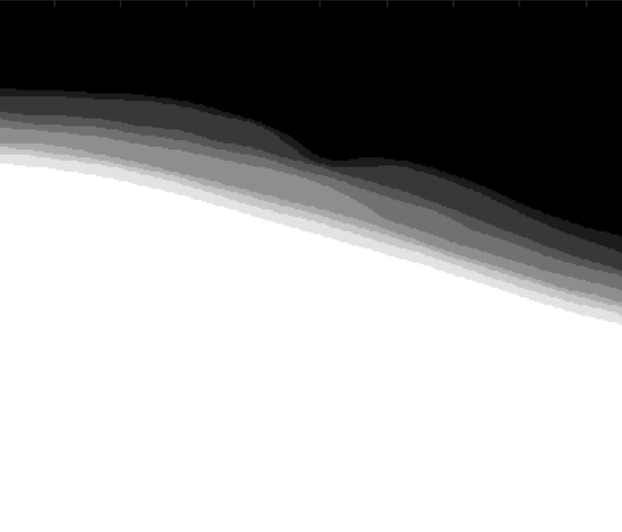} &
\includegraphics[height=2.75 cm]{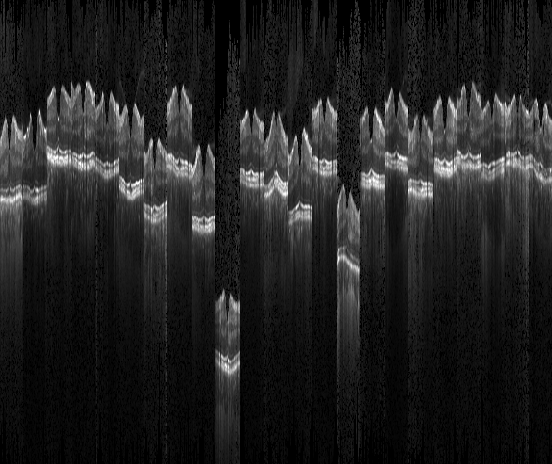} &
\includegraphics[height=2.75 cm]{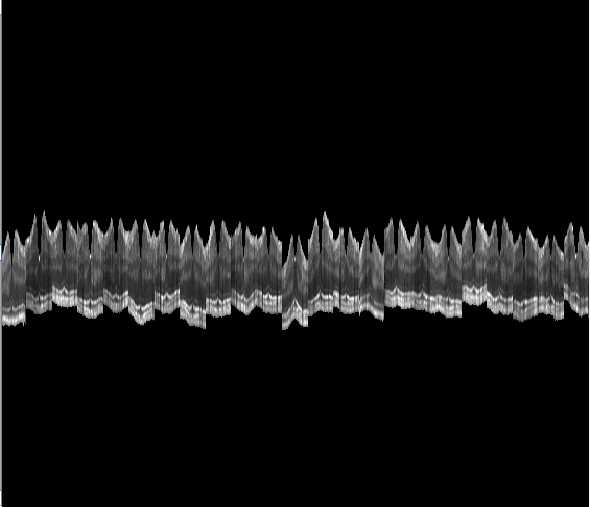} \\
\end{tabular}
 \caption{JHopkins dataset.
 From left to right, fovea-crossing B-scan and corresponding segmentation of the retinal layers (grey scale), vitreous (black), and choroid (white) in one subject; B-scans from the training set in the plane orthogonal to the B-scans; and
 B-scans from the training set after spatial normalization, also in the orthogonal plane to the B-scans.
 }
\label{fig:JHopkins}
\end{figure*}

The dataset used for the training of our segmentation models (JHopkins) belong to the OCT data resource for Multiple Sclerosis (MS) and Healthy Controls (HC)~\cite{He_19}.
This database is made of 35 OCT volumes from a Spectralis Scanner under FastMac protocol.
It is provided with the manual delineations of nine retinal boundaries, or equivalently eight retinal regions, detailed in  Table~\ref{table:RetinalRegions}.
The cohort of subjects contains 14 healthy controls and 21 multiple sclerosis patients with age and gender information.
Volumes consist of 49 equally spaced macular BScans of size $496 \times 1024$.
Figure~\ref{fig:JHopkins} shows an example of a fovea-crossing B-scan and the corresponding segmentation of the retinal regions.
In addition, the figure shows the stacked B-scans of the training set in the orthogonal plane before and after spatial normalization.


\begin{table}
\caption{Anatomical Retinal regions with standard abbreviations of the JHopkins retinal regions.}
\label{table:RetinalRegions}
\centering
\begin{tabular}{|c|c|}
\hline
Name & Abbreviation \\
\hline
\hline
Retinal Nerve Fiber Layer & RNFL \\
Ganglion Cell Layer + Inner Plexiform Layer & GCL+IPL \\
Inner Nuclear Layer & INL \\
Outer Plexiform Layer & OPL \\
Outer Nuclear Layer & ONL \\
Inner photoreceptor segments & IS \\
Outer photoreceptor segments & OS \\
Retinal Pigment Epithelium & RPE \\
\hline
\end{tabular}
\end{table}

\subsubsection{Miguel Servet datasets}

The datasets used for the evaluation of our segmentation models (MServet) were collected during routine clinical practice at
Miguel Servet Hospital, Spain.
The acquisition covers an extensive 10-year period.
The datasets include healthy control patients (HC), samples from the most prevalent neurodegenerative diseases
(Multiple Sclerosis (MS), Parkinson’s (PK), and Alzheimer’s (AD)), and Steinert’s disease (St).
These subjects were recruited by three specialist clinicians: one ophthalmologist specialised in neuro-ophthalmology,
one neurologist specialised in demyelinating diseases, and one neurologist specialist in movement disorders and dementia.
These datasets are intended to be used in studies aimed at characterizing the impact of these diseases on the retina.
The exclusion criteria employed were best corrected visual acuity lower than 0.5 using Snellen charts, refractive errors higher
than 5 dioptres of spherical equivalent refraction or 3 dioptres of astigmatism, intraocular pressure higher than 20 mmHg, media opacifications (nuclear colour/opalescence, cortical or posterior subcapsular lens opacity $>$ 2, according to the Lens Opacities Classification System III), abnormal ocular findings, concomitant ocular diseases such as glaucoma or retinal pathology, and other
systemic conditions that could affect the visual system.
All procedures are in accordance with the Declaration of Helsinki.
Written informed consent was collected from all participants in the study, and the experimental protocol was approved by the Ethics Committee of the Miguel Servet Hospital (CEICA, registration number C.I. PI21/113).


In FastMac protocol, volumes consist of 25 equally spaced macular BScans of size $496 \times 512$.
In PPole protocol, volumes consist of 61 equally spaced macular BScans of size $496 \times 765$.
Images were subject to quality control, and scans failing to meet predefined acquisition quality criteria were excluded.
Table~\ref{table:MiguelServetPopulation} gathers the number of subjects and the number of images used in this study.



\begin{table}
\caption{Population of MServet datasets. Number of subjects and number of images (left/right eye). }
\label{table:MiguelServetPopulation}
\centering
\begin{tabular}{|c|c|c|}
\hline
Group & FastMac & PPole \\
\hline
\hline
Healthy Control (HC) & 75: 101/103 & 64: 64 / 64 \\
Multiple Sclerosis (MS) & 152: 335 / 334 & 29: 34 / 36 \\
Parkinson (PK) & 82: 116 / 121 & 0 \\
Alzheimer (AD) & 23: 21 / 23 & 0 \\
Steinert (St) & 0 & 71: 68 / 71 \\
\hline
\end{tabular}
\end{table}

\subsection{Data preprocessing methods}

\subsubsection{Data augmentation}

Data augmentation is a preprocessing incorporated during training as a standard strategy to enhance the generalization capability of learning-based models.
In this work, the applied augmentations can be categorized into spatial parametric transformations and intensity transformations.
The considered spatial transformations include horizontal flipping and low-angle rotations.
These transformations are applied jointly to both the B-scan images and their corresponding segmentation masks, ensuring spatial consistency.

The considered intensity transformations involve photometric modifications such as Gaussian noise, Gaussian blur, brightness and contrast adjustments, and gamma correction.
These transformations mimic photometric variability typically observed in OCT datasets.
In this case, augmentations are applied exclusively to the B-scan images.
All transformations are applied probabilistically, such that not all images undergo the same modifications, thereby increasing the variability of the training data.

For MGU-Net and SD-LayerNet, the augmentation pipeline was implemented using the Albumentations library~\cite{Buslaev_2020}.
In the case of nnU-Net, similar augmentation strategies are implemented internally, and comparable augmentation behavior was ensured through its configuration.


\subsubsection{Spatial normalization}


Spatial normalization is applied to the OCT volumetric images before training.
We propose the following procedure.
First, the 25th B-scan (JHopkins) or the 13th B-scan (MServet) was selected for the FastMac protocol and the 31st B-scan (MServet) for the PPole protocol as central B-scans.
Then, a coarse retinal segmentation was obtained from the selected slice, including the detection of the Internal Limiting Membrane (ILM) and the Retinal Pigment Epithelium (RPE).
The foveal point (FP) was identified as the location on the ILM with the minimum retinal thickness, defined as the shortest distance between the ILM and the RPE along the A-scans.

The FP was vertically projected onto the RPE curve, and a small segment of the RPE curve around this point was fitted using least squares approximation.
The FP was used as the center of the 3D volume, and the fitted segment was used to estimate the rotation angle that roughly aligns the RPE in the central B-scan with the horizontal axis.
All slices in the 3D volume were rotated using this angle to standardize the alignment and orientation of all B-scans.
Finally, the normalized images were cropped to ensure consistent content across subjects.
This step removed regions near the horizontal boundaries where rotation artifacts from zero-padding or protocol-dependent anatomical differences (such as the presence or absence of the
optic nerve head) could otherwise introduce domain shifts between the training and testing datasets.
Table~\ref{table:DimRed} shows the OCT dimensions before and after consistency cropping.
The normalization pipeline was implemented by adapting He et al. Matlab 2019 tools~\cite{He_19} to our specific datasets.

%
%

\begin{table}
\caption{OCT dimensions before and after consistency cropping.
}
\label{table:DimRed}
\centering
\begin{tabular}{|c|c|c|}
\hline
Dataset & Original & Normalized \\
\hline
\hline
JHopkins & 496 x 1024 x 49 & 496 x 900 x 49 \\
FastMac & 496 x 512 x 25 & 496 x 400 x 25 \\
PPole & 496 x 765 x 61 & 496 x 600 x 61 \\
\hline
\end{tabular}
\end{table}

\subsection{Retinal layer segmentation methods}

\subsubsection{nnU-Net}

nnU-Net is a self-configuring framework built upon the UNet architecture for medical image segmentation.
As implied by its name, no-new-UNet, it does not introduce a novel network design, it retains the canonical UNet encoder-decoder structure and does not introduce deep architecture modifications.
However, its core contributions in the pipeline design, configuration, and training substantially improve the performance over the original U-Net,
making nnU-Net a highly competitive baseline for medical image segmentation.

The central feature of nnU-Net is its fully automated pipeline configuration.
This allows the method to handle a wide range of input data with varying dimensions (2D/3D), sizes, and resolutions, providing a robust starting point without any manual tuning.
In addition, nnU-Net includes specific medical image preprocessing and robust training strategies.
The pipeline incorporates the possibility of extensive data augmentation in spatial (rigid/non-rigid), intensity, and gamma domains.
These components improve training stability and generalization, particularly on small and heterogeneous datasets commonly encountered in medical imaging.

Furthermore, nnU-Net includes dataset-adaptive post-processing and model selection mechanisms.
These strategies further enhance segmentation reliability and consistently outperform a single U-Net trained without such comprehensive automation.

In this work, we used the publicly available code from the nnU-Net repository (https://github.com/MIC-DKFZ/nnUNet).
nnU-Net provides an intuitive, ready-to-run framework for segmentation.
However, its complex pipeline can be difficult to customize, and most users interact with it as a fully configurable black box without the flexibility to modify individual components outside the provided training scripts. This can limit its adaptability for specialized tasks or integration with alternative preprocessing, augmentation, or loss strategies.
For this reason, we employed the scripts as provided in our study, treating nnU-Net as a black-box baseline method.
Although nnU-Net does not rely on Albumentations, it uses a conceptually similar online data augmentation strategy
implemented using its own framework. The choice of transformations and parameter ranges ensured an augmentation
behavior comparable with Albumentations.

\subsubsection{MGU-Net}

MGU-Net is a segmentation architecture specifically designed for OCT images acquired in the peripapillary region~\cite{Li_21}.
The method aims to identify both the optic disc and the retinal layers surrounding the optic nerve head.
Although MGU-Net is based on the UNet architecture, it introduces several improvements to address the specific challenges of this type of imaging, such as poorly defined boundaries, anatomical variability, and artifacts.
As its name suggests, Multi-scale Graph Convolutional Network-assisted UNet incorporates a multi-scale graph convolutional network (GCN),
enabling the model to capture broad spatial relationships between different anatomical regions at different scales.
This network is embedded in a Multi-scale Global Reasoning Module (MGRM), positioned between the encoder and the decoder, acting as a bridge between the two.
By combining these components, the method integrates global context with local details, handles the variability in retinal layer thickness, and favours the preservation of the relative positions of the layers.
This capability is particularly valuable for retinal segmentation, where the layers have a fixed anatomical order.
While MGU-Net was originally developed for the peripapillary region, it can also be adapted to the macular region~\cite{Gende_23}.

We used the publicly available code from the MGU-Net repository (https://github.com/Jiaxuan-Li/MGUNet).
We adapted the original peripapillary implementation to the macular procedure described above.
During training, we introduced a delay between the training of odsMGU-Net and the subsequent incorporation of lsMGU-Net, otherwise, we experienced convergence problems.


\subsubsection{SD-LayerNet}

SD-LayerNet was initially introduced in~\cite{Fazekas_22} and later extended in SD-LayerNet2~\cite{Fazekas_25} as a semi-supervised retinal layer segmentation method for OCT images.
SD-LayerNet formulation is inspired by SDNet~\cite{Chartsias_19}, which proposed a segmentation strategy based on disentangled representations separating spatial anatomical factors from non-spatial image modality factors.
SDNet incorporates an unsupervised branch driven by an autoencoder trained to learn a reconstruction of the original image using the predicted segmentation and the disentangled factors.
SD-LayerNet integrates SDNet ideas with anatomical priors specific to retinal structure such as typical layer positions, topological consistency, curvature constraints, and additional shape-based regularizers.
These priors guide the supervised segmentation module. The autoencoder is conditioned on the layer predictions generated by the supervised module, allowing both components to reinforce each other during training based on segmentation and reconstruction quality.

Two stages can be observed during the training process.
During the transient phase, the network progressively learns the retinal layers one after another, which gives the method its name.
During the stationary phase, the model is refined, allowing the supervised and unsupervised losses to stabilize and converge toward improved segmentation performance.
For this work, we adapted the publicly available code from the SD-LayerNet repository (https://github.com/ABotond/SDLayerNet).
It should be noted that the repository does not include the training and inference pipelines.
In particular, several components of the network require additional inputs derived from both the image and the segmentation, which are not documented or implemented in the original release.
As a result, a substantial reverse-engineering effort was needed to reconstruct the missing training logic and ensure the full architecture operated as intended.
In addition, we extended the original implementation, which was constrained to $128 \times 128$ images, to support rectangular higher-resolution OCT images.


The implementation details are provided in the Supplementary Material.

\section{Results}
\label{sec:Results}


\subsection{Evaluation metrics}

In this work, we evaluate segmentation performance at three complementary levels: B-scan level using overlap-based metrics for each layer,
A-scan level using topology-based metrics, and en-face level through thickness- and surface-based measures.


%
%
%
%
%

\subsubsection{Dice Similarity Coefficient}

The Dice Similarity Coefficient (DSC) is a region-based evaluation metric widely used in image segmentation and non-rigid registration tasks.
DSC measures the overlap between predicted and ground truth regions.
Given two binary representations of the object of interest, $X$ (predicted) and $Y$ (ground truth), the DSC is defined as
\begin{equation}
 DSC = \frac{2 \lvert X \cap Y \rvert}{\lvert X \rvert + \lvert Y \rvert}.
\end{equation}

\noindent In OCT imaging, DSC is commonly computed at the B-scan level for each retinal layer, providing a layer-wise assessment of segmentation accuracy.

\subsubsection{Topological Violations}

At the tissue level, retinal layers form continuous surfaces that preserve a global axial order along the A-scan direction.
Consequently, segmentation surfaces are expected to be topologically equivalent to a plane, i.e., without holes, splits, or self-intersections.
Anatomically implausible configurations such as spurious layer interruptions, unintended fusions, or branching structures are therefore indicative of segmentation failures.
These errors may arise from limitations of the segmentation model or from severe image degradations and acquisition artifacts.
This motivates the use of topology-based evaluation metrics that are independent of conventional overlap-based measures (e.g., DSC) or thickness-based errors, 
and that directly assess the structural plausibility of the resulting segmentations.
An additional advantage of these topology-based metrics is that they do not require ground truth annotations.
Instead, they rely solely on fundamental anatomical constraints, enabling their application in realistic clinical scenarios where dense expert annotations are unavailable or impractical to obtain.

Topological violations can be defined at both the A-scan and B-scan levels.
In this work, we focus on violations at the A-scan level, as B-scan-level inconsistencies can be inferred from the accumulation of local A-scan violations, while being more difficult to detect and localize in a reliable manner.
For each A-scan, we quantify the following types of topological violations:

\begin{enumerate}
 \item Absent layers: one or more retinal layers are missing along the A-scan, indicating discontinuities or collapsed structures.
 \item Label duplication: a retinal layer appears more than once along the same A-scan, suggesting spurious splits or branching.
 \item Order violations (flips): the axial ordering of layers is incorrect, violating known anatomical constraints.
 \item Noise violations: isolated or small spurious regions that do not correspond to valid anatomical layers.
\end{enumerate}

\noindent Whenever any of these violations is detected at a given A-scan, the corresponding location is marked as exhibiting a topological inconsistency.
Aggregating these violations across A-scans and subjects provides a quantitative assessment of the structural reliability of the segmentation methods.


\subsubsection{Retinal layer thickness maps}

At the en-face level, segmentation quality can be evaluated through the analysis of retinal layer thickness maps, which provide a global and clinically meaningful representation of the retinal structure.
Thickness maps are obtained by computing the distance between consecutive layer boundaries along each A-scan. 
To quantify discrepancies between predicted and ground truth thickness, we employ thickness-based error metrics such as the mean absolute difference (MAD) between the estimated $S$ and the ground truth $\hat{S}$
surfaces for each layer

\begin{equation}
\mathrm{MAD} = \frac{1}{N} \sum_{i=1}^{N} \left| S(i) - \hat{S}(i) \right|,
\end{equation}

\noindent and the squared difference (SD) between the estimated $T$ and the ground truth $\hat{T}$ thickness

\begin{equation}
\mathrm{SD} = \frac{1}{N} \sum_{i=1}^{N} \left( T(i) - \hat{T}(i) \right)^2.
\end{equation}

\noindent These metrics capture deviations in layer thickness and provide a complementary perspective to overlap-based measures, being particularly sensitive to boundary inaccuracies and cumulative segmentation errors.

In addition to quantitative error measures, thickness maps allow the visual identification of structural anomalies that are indicative of segmentation failures.
Typical examples include noise, regions with near-zero thickness or abnormally large thickness values, abrupt lateral variations, or persistent vessel-like structures in avascular layers which are inconsistent with known anatomical
patterns.
These irregularities are particularly relevant from a clinical perspective, as they may affect the identification of retinal biomarkers.
Even when considered qualitatively, thickness maps provide a highly informative representation, enabling clear visualization of segmentation defects across individual retinal layers.
Thickness-derived representations are directly related to clinically used measurements and offer an intuitive means to assess whether a given segmentation is sufficiently accurate for reliable downstream analysis
such as biomarker extraction and disease characterization.

\subsection{Results in JHopkins test set}


\subsubsection{Dice Similarity Coefficient}

\begin{table*}[t]
\centering
\tiny
\caption{JHopkins test dataset. Quantitative results on the normalized OCT images.
Mean and standard deviation of the DSC values in the retinal layers.
Upper table, models trained with original training set.
Lower table, models trained with normalized training set.
}
\label{table:JHopkins_DSC}
\begin{tabular}{c}
Original training set \\
\begin{tabular}{|c|ccccccccc|}
\hline
Method & RNFL & GCL+IPL & INL & OPL & ONL & IS & OS & RPE & Total $\uparrow$ \\
\hline
\hline
nnU-Net & $ 32.46 \pm 11.13 $ & $ 54.04 \pm 27.47 $ & $ 65.33 \pm 9.20 $ & $ 81.82 \pm 5.66 $ & $ 84.23 \pm 7.04 $ & $ 74.54 \pm 6.89 $ & $ 46.50 \pm 27.08 $ & $ 57.20 \pm 10.72 $ & $ 62.01 \pm 22.89 $ \\
MGU-Net & $ 84.51 \pm 7.70 $ & $ 84.64 \pm 5.48 $ & $ 64.03 \pm 11.83 $ & $ 73.65 \pm 8.69 $ & $ 88.11 \pm 5.71 $ & $ 76.29 \pm 8.40 $ & $ 78.95 \pm 5.56 $ & $ 86.39 \pm 3.72 $ & $ 79.57 \pm 10.66 $ \\
SD-LayerNet & $ \mathbf{90.95 \pm 3.80 }$ & $ \mathbf{94.01 \pm 3.28} $ & $ 87.05 \pm 3.66 $ & $ 89.66 \pm 2.39 $ & $ 94.68 \pm 1.78 $ & $ 86.48 \pm 3.49 $ & $ 87.79 \pm 3.65 $ & $ 92.41 \pm 2.49 $ & $ \mathbf{90.38 \pm 4.31} $ \\
\hline
nnU-Net aug & $ 25.84 \pm 18.22 $ & $ 16.40 \pm 21.06 $ & $ 27.56 \pm 30.96 $ & $ 31.26 \pm 32.68 $ & $ 28.34 \pm 28.22 $ & $ 40.35 \pm 25.72 $ & $ 68.17 \pm 12.42 $ & $ 23.70 \pm 12.66 $ & $ 32.70 \pm 28.11 $ \\
MGU-Net aug & $ \mathbf{91.14 \pm 4.58} $ & $ \mathbf{ 94.59 \pm 2.67} $ & $ 86.67 \pm 3.27 $ & $ 89.76 \pm 2.37 $ & $ 94.47 \pm 1.78 $ & $ 86.55 \pm 3.47 $ & $ 87.99 \pm 3.36 $ & $ 93.50 \pm 1.92 $ & $\mathbf{ 90.58 \pm 4.38 }$ \\
SD-LayerNet aug & $ 90.78 \pm 4.73 $ & $ 94.40 \pm 2.58 $ & $ 86.39 \pm 3.53 $ & $ 89.12 \pm 2.56 $ & $ 94.61 \pm 1.55 $ & $ 87.21 \pm 2.93 $ & $ 87.77 \pm 3.77 $ & $ 92.27 \pm 2.67 $ & $ 90.32 \pm 4.36 $ \\
\hline
\end{tabular}
\\
Normalized training set \\
\begin{tabular}{|c|ccccccccc|}
\hline
Method & RNFL & GCL+IPL & INL & OPL & ONL & IS & OS & RPE & Total $\uparrow$ \\
\hline
\hline
nnU-Net & $ 41.35 \pm 12.14 $ & $ 75.38 \pm 14.91 $ & $ 63.46 \pm 18.22 $ & $ 85.65 \pm 4.41 $ & $ 86.05 \pm 6.71 $ & $ 76.41 \pm 7.48 $ & $ 77.72 \pm 12.45 $ & $ 59.68 \pm 5.36 $ & $ 70.71 \pm 18.03 $ \\
MGU-Net & $ 90.16 \pm 5.16 $ & $ 93.66 \pm 3.27 $ & $ 85.78 \pm 4.93 $ & $ 89.21 \pm 2.96 $ & $ 94.36 \pm 1.78 $ & $ 87.10 \pm 3.55 $ & $ 87.77 \pm 3.47 $ & $ 93.21 \pm 1.98 $ & $ 90.16 \pm 4.69 $ \\
SD-LayerNet & $ \mathbf{91.62 \pm 4.14} $ & $ \mathbf{94.88 \pm 2.37} $ & $ 87.69 \pm 3.38 $ & $ 90.34 \pm 2.32 $ & $ 94.89 \pm 1.63 $ & $ 87.47 \pm 3.21 $ & $ 87.55 \pm 4.55 $ & $ 92.77 \pm 3.08 $ & $\mathbf{ 90.90 \pm 4.36 }$ \\
\hline
nnU-Net aug & $ \mathbf{92.19 \pm 4.46} $ & $ \mathbf{94.93 \pm 2.30} $ & $ 86.68 \pm 4.34 $ & $ 89.73 \pm 3.34 $ & $ 94.97 \pm 1.74 $ & $ 88.53 \pm 3.09 $ & $ 89.36 \pm 3.44 $ & $ 93.30 \pm 2.38 $ & $ \mathbf{91.21 \pm 4.35} $ \\
MGU-Net aug & $ 89.96 \pm 5.43 $ & $ 93.99 \pm 3.26 $ & $ 87.09 \pm 3.46 $ & $ 89.61 \pm 2.30 $ & $ 94.59 \pm 1.74 $ & $ 87.36 \pm 3.39 $ & $ 88.06 \pm 3.64 $ & $ 93.33 \pm 2.14 $ & $ 90.50 \pm 4.40 $ \\
SD-LayerNet aug & $ 90.03 \pm 5.44 $ & $ 94.09 \pm 3.22 $ & $ 87.25 \pm 3.19 $ & $ 89.88 \pm 2.05 $ & $ 94.65 \pm 1.61 $ & $ 87.09 \pm 3.17 $ & $ 87.81 \pm 3.92 $ & $ 93.23 \pm 2.21 $ & $ 90.50 \pm 4.39 $ \\
\hline
\end{tabular}
\\
\end{tabular}
\end{table*}

Quantitative results of the DSC metrics on the JHopkins test dataset are reported in Table~\ref{table:JHopkins_DSC}.
When models are trained on the original dataset, SD-LayerNet consistently achieves the best overall performance, reaching the highest mean DSC across all retinal layers.
In contrast, nnU-Net exhibits substantially lower accuracy in this setting, particularly in challenging layers such as RNFL and GCL+IPL, highlighting its limited robustness when trained on non-normalized and non-augmented data.
The introduction of data augmentation leads to a performance improvement for all methods, with MGU-Net achieving comparable results
to SD-LayerNet. However, nnU-Net remains considerably less stable under this configuration, showing large variability across layers.


When training is performed on the normalized dataset, all methods benefit from improved performance and stability.
In this setting, SD-LayerNet maintains strong and consistent results, while MGU-Net achieves comparable performance across most layers.
Notably, nnU-Net shows a substantial performance gain compared to the original training setup.

The combination of normalization and data augmentation yields the best overall results.
Under this configuration, nnU-Net achieves the highest overall DSC, outperforming both MGU-Net and SD-LayerNet, and demonstrating a marked sensitivity to data preprocessing.
These results indicate that while SD-LayerNet provides robust performance across different settings, nnU-Net can surpass competing methods only when appropriate normalization and augmentation strategies are applied.

Across all experimental settings, INL, IS, and OS consistently emerge as the most challenging layers, with lower DSC values and higher variability across methods.
This trend is stable regardless of the training configuration, suggesting intrinsic difficulties associated with these structures.
This behavior can be attributed to several factors. First, these layers often exhibit lower contrast and less clearly defined boundaries, making their delineation inherently ambiguous.
Second, their relatively small thickness increases sensitivity to minor localization errors, which can significantly affect overlap-based metrics such as DSC.
Finally, anatomical variability may further contribute to the observed performance degradation.

\subsubsection{Topology preserving}

\begin{table}[t]
\centering
\tiny
 \caption{JHopkins test dataset. Topology violations per A-scan. Best Avg/BScan results are shown in bold.}
 \label{table:JHopkins_TopoViol}

 \begin{tabular}{c}

 Original training set \\

 \begin{tabular}{|c|ccccc|}
 \hline
  Model & absent & duplication & flip & noise & Avg / BScan $\downarrow$ \\
  \hline
  \hline

  nnU-Net & -- & -- & -- & -- & -- \\
  MGU-Net & 2592 & 4182 & 2111 & 6917 & 46.07 \\
  SD-LayerNet & 638 & 0 & 0 & 0 & \textbf{1.86} \\
  \hline
  nnU-Net aug & -- & -- & -- & -- & -- \\
  MGU-Net aug & 1127 & 31 & 0 & 49 & 3.52 \\
  SD-LayerNet aug & 66 & 0 & 0 & 0 & \textbf{0.19} \\

  \hline
 \end{tabular}
\\
Normalized training set
\\
 \begin{tabular}{|c|ccccc|}
 \hline
  Model & absent & duplication & flip & noise & Avg / BScan $\downarrow$ \\
  \hline
  \hline

  nnU-Net & -- & -- & -- & -- & -- \\
  MGU-Net & 1251 & 2068 & 647 & 1504 & 15.95 \\
  SD-LayerNet & 66 & 0 & 0 & 0 & \textbf{0.19} \\
  \hline
  nnU-Net aug & 935 & 103 & 7 & 55 & 3.21 \\
  MGU-Net aug & 647 & 797 & 133 & 658 & 6.52 \\
  SD-LayerNet aug & 257 & 0 & 0 & 0 & \textbf{0.75} \\

  \hline
 \end{tabular}
 \end{tabular}
\end{table}

Table~\ref{table:JHopkins_TopoViol} reports the number of topology violations across all test subjects and BScans.
The vitreous and choroid regions were excluded from the topology analysis, as errors at the RPE boundary often resulted in spurious violations, including noise-like artifacts and large disconnected regions.
These errors are primarily driven by boundary ambiguity and signal variability outside the retinal layers, rather than by the intrinsic topological consistency of the segmentation models.
Their inclusion led to an overestimation of topology violations and hindered a fair comparison across methods.
Therefore, the analysis was restricted to the internal retinal layers to better reflect structural consistency within the region of interest.

Clear differences are observed between architectures.
SD-LayerNet consistently produces nearly topology-preserving segmentations, with zero duplication, flip, and noise errors across all configurations, and a very low number of absent layers.
In contrast, MGU-Net exhibits a significantly higher number of violations, particularly in the original training setting, where all error types are frequently observed.
Data augmentation substantially reduces topology violations for MGU-Net, decreasing the average number of errors per B-scan from 46.07 to 3.52.
A similar improvement is observed when training on normalized data, with a decrease from 15.95 to 6.52.
In comparison, SD-LayerNet maintains stable and near-zero violation rates regardless of the training configuration.
Interestingly, nnU-Net produces topologically consistent results when trained with both normalization and augmentation. 
Overall, these results indicate that SD-LayerNet enforces stronger implicit topological constraints, whereas other methods require normalization and augmentation to reduce structural inconsistencies.

\begin{figure}
\centering
\small
\begin{tabular}{c}
nnU-Net \\
\includegraphics[width=7.0 cm]{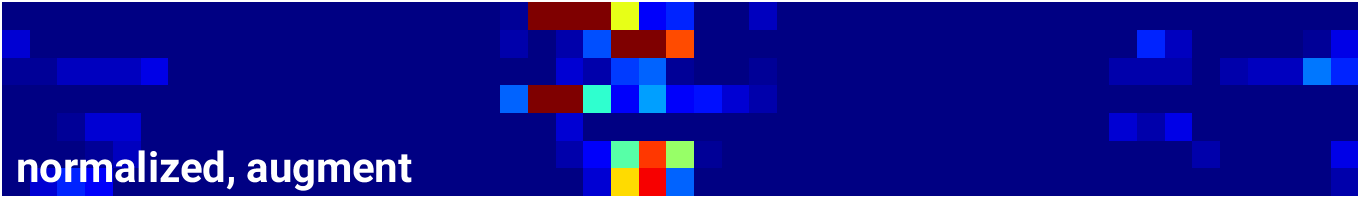} \\
MGU-Net \\
\includegraphics[width=7.0 cm]{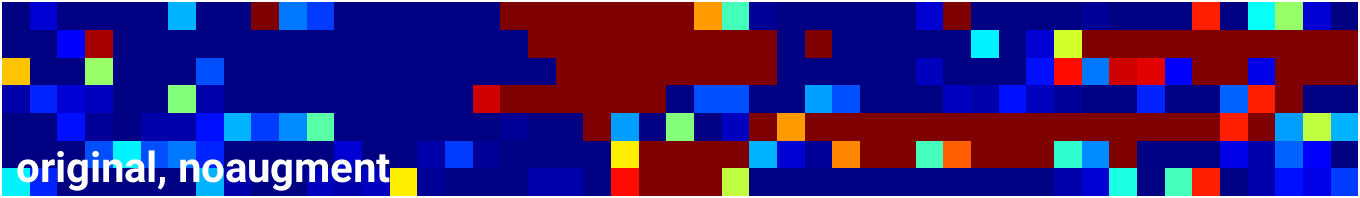} \\
\includegraphics[width=7.0 cm]{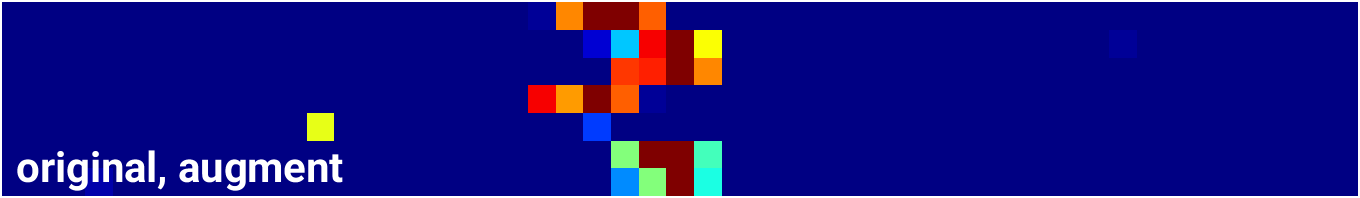} \\
\includegraphics[width=7.0 cm]{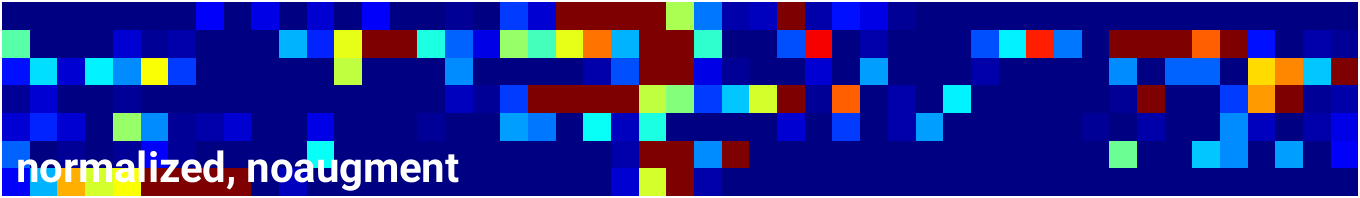} \\
\includegraphics[width=7.0 cm]{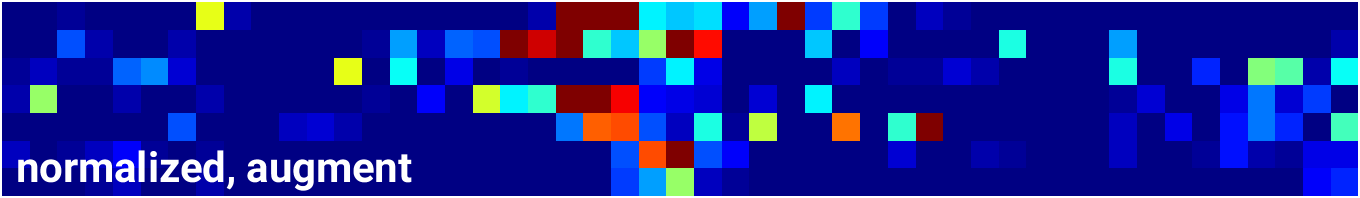} \\
SD-LayerNet \\
\includegraphics[width=7.0 cm]{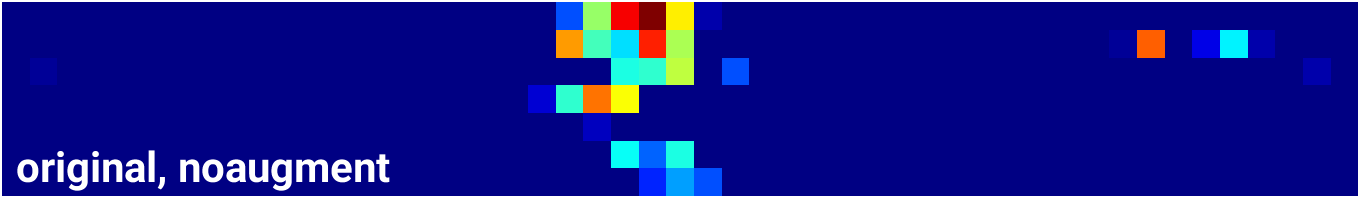} \\
\includegraphics[width=7.0 cm]{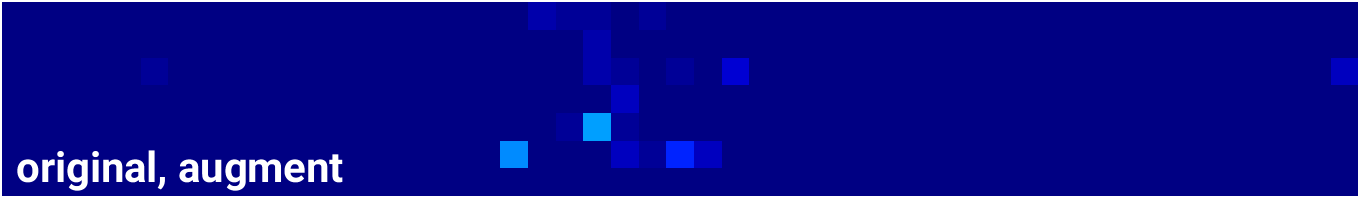} \\
\includegraphics[width=7.0 cm]{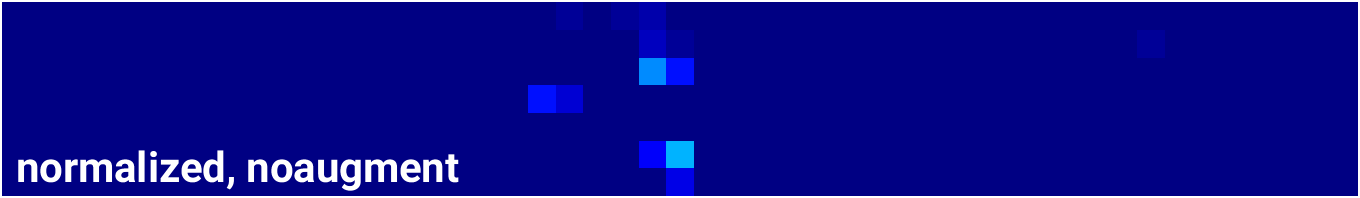} \\
\includegraphics[width=7.0 cm]{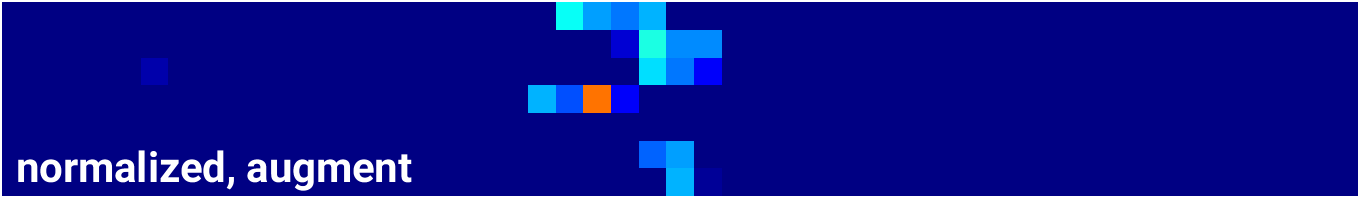} \\

\end{tabular}
\caption{
JHopkins test dataset. Distribution of topology violations per subject (rows) and B-scan (columns).
Each entry represents the total number of quantified topology violations.
Comparison across methods is enabled thanks to the spatially normalized domain.
}
\label{fig:TopoViolsApril}
\end{figure}

Figure~\ref{fig:TopoViolsApril} shows, for each method, the distribution of topology violations per subject and B-scan.
The different experiments are comparable thanks to the use of a spatially normalized test set.
These results visually corroborate the quantitative findings reported in Table~\ref{table:JHopkins_TopoViol}.
SD-LayerNet consistently exhibits the lowest number of violations across all configurations, confirming its superior topological stability.
Across B-scans, topology violations are predominantly concentrated in the central slices.
These correspond to the central macular region, where the presence of the foveal pit causes the retinal layers to converge and become thinner,
increasing the likelihood of structural inconsistencies and segmentation errors.

\subsubsection{En-face thickness analysis}

Table SII in the Supplementary Material reports the mean absolute distance (MAD) errors for all methods across retinal layer boundaries.
Overall, SD-LayerNet consistently achieves lower or comparable errors across most interfaces when trained on the original dataset,
showing improved accuracy and notably reduced variability compared to MGU-Net.
When trained on the original data, MGU-Net exhibits substantially higher errors, particularly in challenging boundaries such as
RNFL-GCL+IPL and INL-OPL, which is consistent with the increased segmentation difficulty observed in these regions.
The introduction of data augmentation leads to a marked reduction in MAD values for MGU-Net, bringing its performance closer to
that of SD-LayerNet across most layers.

Training on the normalized dataset further stabilizes the performance of all methods.
Under this configuration, the gap between MGU-Net and SD-LayerNet is reduced, with both methods achieving similar error levels across most boundaries.
Interestingly, nnU-Net achieves competitive performance when trained with both normalization and augmentation, obtaining the lowest
overall MAD despite its lower robustness in other configurations.

Across all methods, higher errors are consistently observed in interfaces involving inner and outer retinal layers,
such as RNFL-GCL+IPL and INL-OPL, reflecting the increased ambiguity and reduced contrast at these boundaries.
In contrast, more distinct interfaces such as ILM and BM tend to yield lower errors, indicating that boundary definition plays
a key role in segmentation accuracy.

Figure S3 in the Supplementary Material shows the thickness maps of the GCL+IPL layer obtained from the different segmentation methods.
A strong visual agreement can be observed between the maps produced by the best-performing methods and the ground truth, as identified by DSC and topology metrics.
This indicates that these methods are able to preserve not only boundary accuracy but also clinically relevant structural patterns.

At the same time, some methods exhibit localized defects and irregularities in the thickness maps that are not clearly reflected in the topology violation analysis.
These artifacts highlight subtle inconsistencies in the segmentation that are not captured by standard overlap or topological metrics.
Therefore, thickness maps provide a complementary and more sensitive means to assess the structural validity of the segmentation results.

To further quantify these effects, we compute the thickness error as the squared difference between predicted and ground-truth layer thickness.
Table~\ref{table:JHopkins_ThicknessError} reports the thickness error across retinal layers for all methods.
Consistent with previous metrics, clear differences are observed between architectures, particularly when trained on the original dataset.
MGU-Net exhibits substantially higher errors, mainly driven by large deviations in the GCL+IPL and RPE layers, leading to a significantly
higher total error.
In contrast, SD-LayerNet achieves consistently low errors across all layers, indicating a more accurate and stable estimation of retinal thickness.


\begin{table}[t]
\centering
\tiny
\caption{JHopkins test dataset. Thickness errors computed as the squared difference (SD) between predicted
and ground-truth thickness.
}
\label{table:JHopkins_ThicknessError}

\begin{tabular}{c}
Original training set \\
\begin{tabular}{|c|ccccccccc|}
\hline
Method & RNFL & GCL+IPL & INL & OPL & ONL & IS & OS & RPE & Total $\downarrow$ \\
\hline
\hline
MGU-Net & 1.20 & 14.37 & 1.21 & 0.47 & 1.06 & 0.20 & 0.26 & 5.94 & 24.70 \\
SD-LayerNet & \textbf{0.44} & \textbf{0.43} & 0.27 & 0.28 & 0.27 & 0.21 & 0.19 & 0.22 & \textbf{2.31} \\

\hline
MGU-Net aug & \textbf{0.25} & \textbf{0.29} & 0.27 & 0.28 & 0.25 & 0.18 & 0.20 & 0.21 & \textbf{1.93} \\
SD-LayerNet aug & 0.31 & 0.35 & 0.27 & 0.29 & 0.27 & 0.17 & 0.19 & 0.22 & 2.07 \\

\hline
\end{tabular}
\\
Normalized training set \\
\begin{tabular}{|c|ccccccccc|}
\hline
Method & RNFL & GCL+IPL & INL & OPL & ONL & IS & OS & RPE & Total $\downarrow$ \\
\hline
\hline
MGU-Net & 0.33 & 0.35 & 0.34 & 0.33 & 0.30 & 0.18 & 0.27 & 0.72 & 2.81 \\
SD-LayerNet & \textbf{0.23} & \textbf{0.23} & 0.26 & 0.25 & 0.26 & 0.17 & 0.23 & 0.26 & \textbf{1.89} \\
\hline
nnU-Net aug & \textbf{0.18} & \textbf{0.26} & 0.27 & 0.27 & 0.27 & 0.18 & 0.19 & 0.22 & \textbf{1.83} \\
MGU-Net aug & 0.36 & 0.37 & 0.28 & 0.29 & 0.25 & 0.18 & 0.21 & 0.22 & 2.15 \\
SD-LayerNet aug & 0.32 & 0.32 & 0.25 & 0.26 & 0.27 & 0.18 & 0.20 & 0.21 & 2.01 \\

\hline
\end{tabular}
\\
\end{tabular}
\end{table}

The use of data augmentation leads to a pronounced reduction in thickness errors for MGU-Net, bringing its performance closer to that of
SD-LayerNet.
Under this configuration, both methods achieve comparable total errors, although SD-LayerNet maintains slightly more uniform performance
across layers.
Training on the normalized dataset further reduces the gap between methods, with all architectures achieving similar error levels.
In this setting, nnU-Net trained with augmentation yields the lowest overall error, closely followed by SD-LayerNet, while MGU-Net shows slightly higher variability.

Across all configurations, larger errors are consistently observed in layers such as GCL+IPL, which are more sensitive to boundary inaccuracies and local structural inconsistencies.
These results highlight that thickness-based metrics capture complementary aspects of segmentation quality, revealing errors that are not fully reflected by overlap (DSC) or boundary-based (MAD) measures.
In scenarios where ground truth segmentations are not available, thickness map visualization still offers a valuable complementary approach for qualitatively assessing segmentation quality.

\subsection{Results in MServet set}

The MServet datasets reflect acquisition under real clinical conditions and introduce several challenges for our models.
First, adjacent retinal layers may exhibit low contrast and subtle intensity differences in some locations, particularly in the inner retina,
making boundary delineation inherently difficult.
Vascular features can induce local intensity distortion that propagates vertically across layers.
The presence of intrinsic speckle noise degrades layer visibility, affecting thin interfaces.
Additionally, some B-scans exhibit strong shadowing artifacts that obscure a segment of the deeper retinal structures.
The limited axial and lateral resolution also leads to cases where thin layers are not clearly separable, introducing unavoidable structural
ambiguity.

A key challenge for the deep-learning models is the high inter-subject anatomical variability, especially pronounced between protocols,
where layer thickness, curvature, and reflectivity can vary substantially. This variability is further amplified by domain shifts associated
with differences in devices, and population characteristics, leading to changes in intensity distributions.
For this reason, we have included the semi-supervised module of SD-LayerNet in the evaluation.

\subsubsection{MServet. FastMac protocol.}

Table~\ref{table:MServetFM_Topo_Viol} reports the number of topology violations on the MServet dataset under the FastMac protocol.
Unlike the controlled JHopkins test set, this dataset reflects real clinical conditions, including subjects with neurodegenerative diseases,
which introduces higher anatomical variability and increased segmentation difficulty.
As a result, a larger number of violations is observed across all methods and configurations.

The already observed differences between architectures remain evident. SD-LayerNet consistently produces topology-preserving segmentations,
with zero duplication, flip, and noise errors across all settings. The number of absent layers is also significantly lower compared to other
methods, particularly when normalization and data augmentation are combined. The semi-supervised variant further improves performance for
specific configurations, but only-supervised SD-LayerNet consistently achieves the lowest average number of violations per B-scan across almost all diagnostic groups.

In contrast, MGU-Net exhibits a high number of violations, especially in configurations without augmentation, where all error types are present.
Although the introduction of data augmentation reduces these errors, the number of violations remains substantially higher than for SD-LayerNet.
This trend is consistent across all pathologies, indicating limited robustness to anatomical variability.

A particularly notable behavior is observed for nnU-Net, where a dramatic increase in topology violations is found.
This suggests that, under strong domain shifts and complex clinical data, nnU-Net may become unstable and generate structurally inconsistent segmentations.
This contrasts with the behavior observed in the JHopkins dataset, highlighting the importance of evaluating models under realistic clinical conditions.

Across diagnostic groups, topology violations are consistently higher in pathological cases, particularly in Parkinson’s and Alzheimer’s disease,
where structural alterations and increased variability further challenge the segmentation task. Despite this, SD-LayerNet maintains stable performance,
confirming its ability to enforce strong implicit topological constraints even in the presence of disease-related variability.
Overall, these results reinforce the findings observed on the JHopkins dataset, demonstrating that SD-LayerNet provides superior topological consistency and robustness,
while other architectures require additional strategies and still struggle to maintain structural coherence under realistic clinical conditions.

\begin{table}[t]
\centering
\tiny
 \caption{MServet FastMac datasets. Topology violations per A-scan. Best results are shown in bold.}
 \label{table:MServetFM_Topo_Viol}

 \begin{tabular}{c}

 Healthy Controls \\

 \begin{tabular}{|c|ccccc|}
 \hline
  Model & absent & duplication & flip & noise & Avg / BScan $\downarrow$ \\
  \hline
  \hline
  SD-LayerNet orig & 34731 & 0 & 0 & 0 & 6.81 \\
  SD-LayerNet semis orig & 9235 & 0 & 0 & 0 & \textbf{1.81} \\
  \hline
  MGU-Net orig + aug & 21724 & 3636 & 79 & 1452 & 5.27 \\
  SD-LayerNet orig + aug & 6320 & 0 & 0 & 0 & \textbf{1.24} \\
  SD-LayerNet semis orig + aug & 8966 & 0 & 0 & 0 & 1.76\\
  \hline
  \hline
  SD-LayerNet norm & 8603 & 0 & 0 & 0 & 1.69 \\
  SD-LayerNet semis norm & 3584 & 0 & 0 & 0 & \textbf{0.70} \\
  \hline
  nnU-Net norm + aug & 78905 & 266550 & 16029 & 51593 & 70.61 \\
  MGU-Net norm + aug & 18585 & 62674 & 6226 & 31836 & 23.40 \\
  SD-LayerNet norm + aug & 4549 & 0 & 0 & 0 & \textbf{0.89} \\
  SD-LayerNet semis norm + aug & 8004 & 0 & 0 & 0 & 1.57 \\
  \hline
 \end{tabular}

 \\
 Multiple Sclerosis \\

 \begin{tabular}{|c|ccccc|}
 \hline
  Model & absent & duplication & flip & noise & Avg / BScan $\downarrow$ \\
  \hline
  \hline
  SD-LayerNet orig & 356538 & 0 & 0 & 0 & 21.32 \\
  SD-LayerNet semis orig & 88637 & 0 & 0 & 0 & \textbf{5.30} \\
  \hline
  MGU-Net orig + aug & 114455 & 18836 & 907 & 7245 & 8.46 \\
  SD-LayerNet orig + aug & 48389 & 0 & 0 & 0 & \textbf{2.89} \\
  SD-LayerNet semis orig + aug & 64855 & 0 & 0 & 0 & 3.88 \\
  \hline
  \hline
  SD-LayerNet norm & 95105 & 0 & 0 & 0 & 5.69 \\
  SD-LayerNet semis norm & 39060 & 0 & 0 & 0 & \textbf{2.34} \\
  \hline
  nnU-Net norm + aug & 216830 & 692362 & 53462 & 161414 & 67.21 \\
  MGU-Net norm + aug & 93703 & 231518 & 28668 & 124182 & 28.58 \\
  SD-LayerNet norm + aug & 21834 & 0 & 0 & 0 & \textbf{1.31} \\
  SD-LayerNet semis norm + aug & 55795 & 0 & 0 & 0 & 3.34 \\
  \hline
 \end{tabular}

 \\

 Parkinson's Disease \\

 \begin{tabular}{|c|ccccc|}
 \hline
  Model & absent & duplication & flip & noise & Avg / BScan \\
  \hline
  \hline
  SD-LayerNet orig & 153671 & 0 & 0 & 0 & 25.94 \\
  SD-LayerNet semis orig & 42528 & 0 & 0 & 0 & \textbf{7.18} \\
  \hline
  MGU-Net orig + aug & 46466 & 6939 & 260 & 2644 & 9.50 \\
  SD-LayerNet orig + aug & 14500 & 0 & 0 & 0 & \textbf{2.45} \\
  SD-LayerNet semis orig + aug & 21078 & 0 & 0 & 0 & 3.56 \\
  \hline
  \hline
  SD-LayerNet norm & 40513 & 0 & 0 & 0 & 6.84 \\
  SD-LayerNet semis norm & 17590 & 0 & 0 & 0 & \textbf{2.97} \\
  \hline
  nnU-Net norm + aug & 67205 & 373887 & 20855 & 74849 & 90.60 \\
  MGU-Net norm + aug & 41623 & 81314 & 9870 & 46185 & 30.21 \\
  SD-LayerNet norm + aug & 8635 & 0 & 0 & 0 & \textbf{1.46} \\
  SD-LayerNet semis norm + aug & 16029 & 0 & 0 & 0 & 2.71 \\
  \hline
 \end{tabular}

  \\

 Alzheimer's Disease \\

 \begin{tabular}{|c|ccccc|}
 \hline
  Model & absent & duplication & flip & noise & Avg / BScan \\
  \hline
  \hline
  SD-LayerNet orig & 67505 & 0 & 0 & 0 & 61.37 \\
  SD-LayerNet semis orig & 16182 & 0 & 0 & 0 & \textbf{14.71} \\
  \hline
  MGU-Net orig + aug & 17650 & 715 & 34 & 816 & 17.47 \\
  SD-LayerNet orig + aug & 2424 & 0 & 0 & 0 & \textbf{2.20} \\
  SD-LayerNet semis orig + aug & 5848 & 0 & 0 & 0 & 5.32 \\
  \hline
  \hline
  SD-LayerNet norm & 15959 & 0 & 0 & 0 & 14.51 \\
  SD-LayerNet semis norm & 4168 & 0 & 0 & 0 & \textbf{3.79} \\
  \hline
  nnU-Net norm + aug & 24453 & 95222 & 8259 & 21644 & 135.98 \\
  MGU-Net norm + aug & 12322 & 21991 & 2221 & 13433 & 45.42 \\
  SD-LayerNet norm + aug & 2123 & 0 & 0 & 0 & \textbf{1.93} \\
  SD-LayerNet semis norm + aug & 4900 & 0 & 0 & 0 & 4.45 \\
  \hline
 \end{tabular}

 \\
 \end{tabular}
\end{table}

\subsubsection{MServet. PPole protocol.}

Table~\ref{table:MServetPP_Topo_Viol} reports topology violations on the MServet dataset under the PPole protocol.
Compared to the FastMac setting, the PPole acquisition introduces increased spatial variability due to its wider coverage,
which further challenges segmentation consistency. This is reflected in the higher number of violations observed across most methods.
Despite this increased complexity, the overall trends remain consistent. SD-LayerNet continues to produce topology-preserving segmentations
across all configurations, with zero duplication, flip, and noise errors, and the lowest number of absent layers.
In contrast, MGU-Net and nnU-Net exhibit a substantial degradation in performance, where large numbers of structural errors of all kind emerge.
Across all diagnostic groups, the semi-supervised variants and the use of normalization and augmentation lead to improved results,
although their impact is clearly model-dependent.
In particular, the semi-supervised variant of SD-LayerNet consistently achieves competitive or improved performance, particularly in challenging
configurations without augmentation, where it significantly reduces the number of absent layers.
However, its relative advantage becomes less pronounced when combined with strong data augmentation, suggesting that both strategies provide
complementary regularization effects.
Overall, these findings confirm that SD-LayerNet provides the most robust and stable behavior
under increased spatial variability, reinforcing its suitability for realistic clinical scenarios beyond FastMac acquisitions.

\begin{table}[t]
\centering
\tiny
 \caption{MServet posterior pole datasets. Topology violations per A-scan. Best Avg/BScan results are shown in bold.}
 \label{table:MServetPP_Topo_Viol}

 \begin{tabular}{c}

 Healthy Controls \\

 \begin{tabular}{|c|ccccc|}
 \hline
  Model & absent & duplication & flip & noise & Avg / BScan $\downarrow$ \\
  \hline
  \hline
  SD-LayerNet orig & 97162 & 0 & 0 & 0 & 12.48 \\
  SD-LayerNet semis orig & 19770 & 0 & 0 & 0 & \textbf{2.54} \\
  \hline
  MGU-Net orig + aug & 44518 & 2157 & 105 & 2254 & 6.30 \\
  SD-LayerNet orig + aug & 13730 & 0 & 0 & 0 & 1.76 \\
  SD-LayerNet semis orig + aug & 8240 & 0 & 0 & 0 & \textbf{1.06} \\
  \hline
  \hline
  SD-LayerNet norm & 31841 & 0 & 0 & 0 & 4.09 \\
  SD-LayerNet semis norm & 15628 & 0 & 0 & 0 & \textbf{2.01} \\
  \hline
  nnU-Net norm + aug & 153260 & 48862 & 7520 & 15580 & 23.43 \\
  MGU-Net norm + aug & 49388 & 79730 & 11441 & 58665 & 25.59 \\
  SD-LayerNet norm + aug & 8328 & 0 & 0 & 0 & \textbf{1.07} \\
  SD-LayerNet semis norm + aug & 12280 & 0 & 0 & 0 & 1.58\\
  \hline
 \end{tabular}

 \\
 Multiple Sclerosis \\

 \begin{tabular}{|c|ccccc|}
 \hline
  Model & absent & duplication & flip & noise & Avg / BScan $\downarrow$ \\
  \hline
  \hline
  SD-LayerNet orig & 53398 & 0 & 0 & 0 & 12.52 \\
  SD-LayerNet semis orig & 28681 & 0 & 0 & 0 & \textbf{6.72} \\
  \hline
  MGU-Net orig + aug & 24885 & 1699 & 67 & 2194 & 6.76 \\
  SD-LayerNet orig + aug & 9588 & 0 & 0 & 0 & 2.25 \\
  SD-LayerNet semis orig + aug & 6530 & 0 & 0 & 0 & \textbf{1.53} \\
  \hline
  \hline
  SD-LayerNet norm & 10539 & 0 & 0 & 0  & 2.47 \\
  SD-LayerNet semis norm & 4654 & 0 & 0 & 0 & \textbf{1.09} \\
  \hline
  nnU-Net norm + aug & 37731 & 25263 & 4094 & 8117 & 17.63 \\
  MGU-Net norm + aug & 26354 & 43391 & 7802 & 34341 & 26.23 \\
  SD-LayerNet norm + aug & 5637 & 0 & 0 & 0 & \textbf{1.32} \\
  SD-LayerNet semis norm + aug & 8203 & 0 & 0 & 0 & 1.92 \\
  \hline
 \end{tabular}
 \\

 Steinert's Disease \\

 \begin{tabular}{|c|ccccc|}
 \hline
  Model & absent & duplication & flip & noise & Avg / BScan $\downarrow$ \\
  \hline
  \hline
  SD-LayerNet orig & 107302 & 0 & 0 & 0 & 12.69 \\
  SD-LayerNet semis orig & 16845 & 0 & 0 & 0 & \textbf{1.99} \\
  \hline
  MGU-Net orig + aug & 21483 & 5852 & 83 & 2448 & 3.53\\
  SD-LayerNet orig + aug & 10607 & 0 & 0 & 0 & 1.25 \\
  SD-LayerNet semis orig + aug & 9318 & 0 & 0 & 0 & \textbf{1.10} \\
  \hline
  \hline
  SD-LayerNet norm & 20399 & 0 & 0 & 0 & 2.41 \\
  SD-LayerNet semis norm & 4922 & 0 & 0 & 0 & \textbf{0.58} \\
  \hline
  nnU-Net norm + aug & 119093 & 109362 & 12982 & 27655 & 31.82 \\
  MGU-Net norm + aug & 28047 & 106910 & 11102 & 66326 & 25.11 \\
  SD-LayerNet norm + aug & 5676 & 0 & 0 & 0 & \textbf{0.67} \\
  SD-LayerNet semis norm + aug & 8999 & 0 & 0 & 0 & 1.06 \\
  \hline
 \end{tabular}
 \\
 \end{tabular}
\end{table}

\begin{figure}
\centering
 \includegraphics[width=4.0 cm]{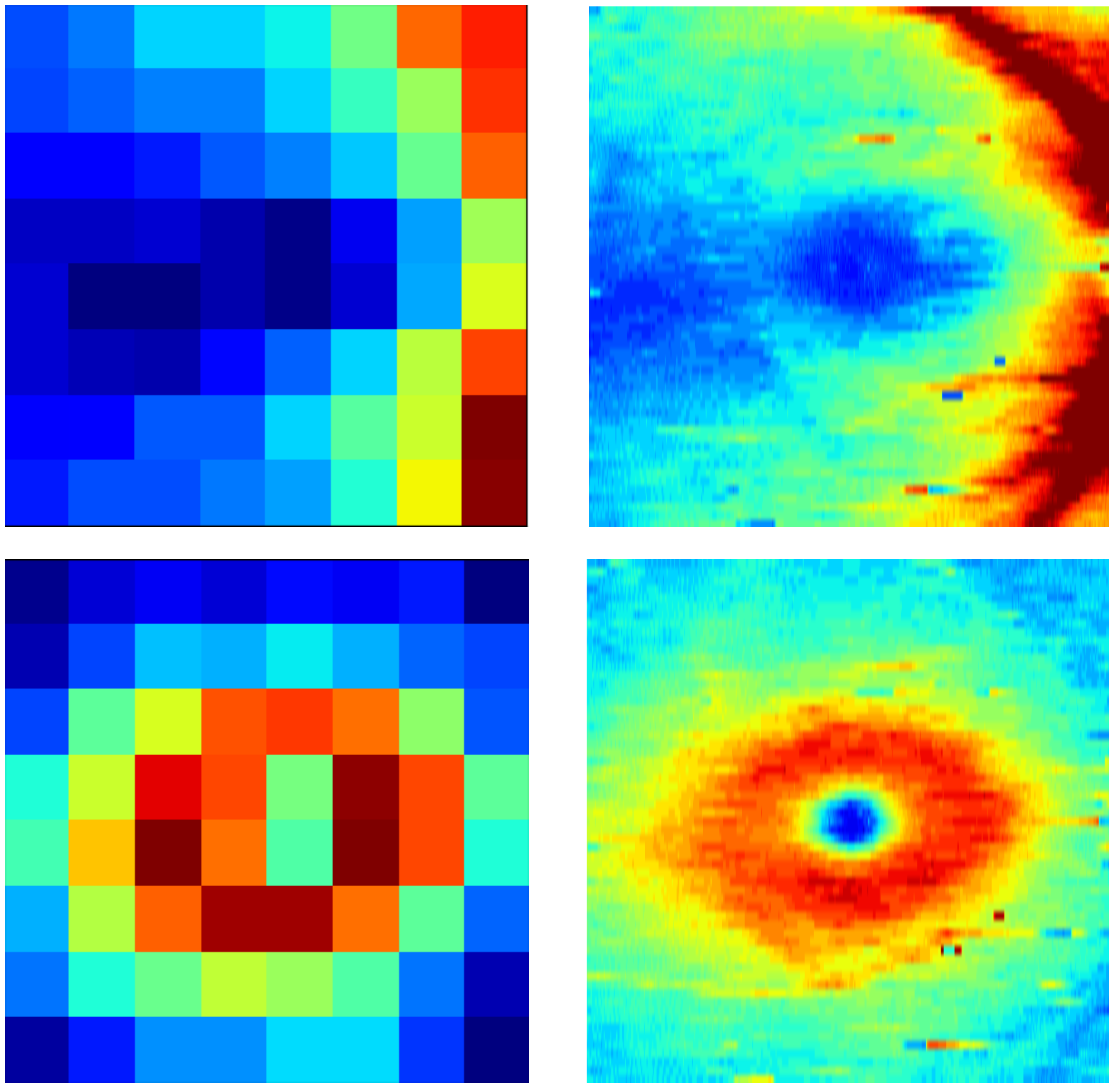} \\
 \vspace{-0.25 cm}
 \caption{MServet PPole datasets. Thickness maps of the RNFL (up) and the GCL+IPL (down) layers provided by Heidelberg
 device (left) and obtained from one of the proposed SD-LayerNet configurations (right).
 }
 \label{fig:HeidelbergVSSD-LayerNet}
\end{figure}

Figure~\ref{fig:HeidelbergVSSD-LayerNet} illustrates a comparison between the thickness maps provided by the Heidelberg device
and those derived from one of the proposed SD-LayerNet configurations. While the Heidelberg maps are limited to a coarse
$8 \times 8$ grid representation, our approach enables the reconstruction of high-resolution thickness maps with substantially
finer spatial detail. This increased resolution allows for a more accurate and intuitive visualization of retinal layer geometry,
facilitating clinical interpretation. Moreover, such dense representations have the potential to provide richer and more informative
input for machine learning and deep-learning methods, potentially improving the sensitivity of AI-based systems for the analysis and
detection of retinal biomarkers.



\subsubsection{MServet. Fast Mac vs PPole qualitative results.}

\begin{figure*}
\centering
\scriptsize
\begin{tabular}{cccccccc}
 \includegraphics[width=2 cm]{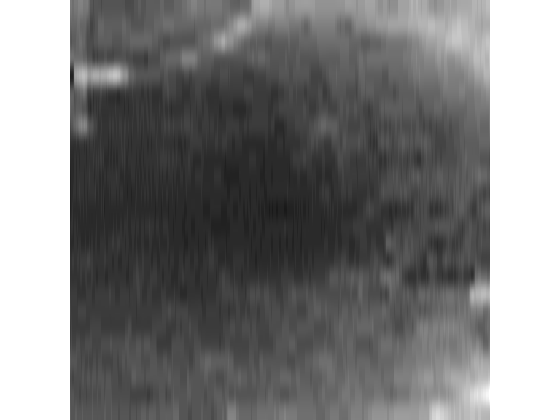} &
 \hspace{-0.5cm}
 \includegraphics[width=2 cm]{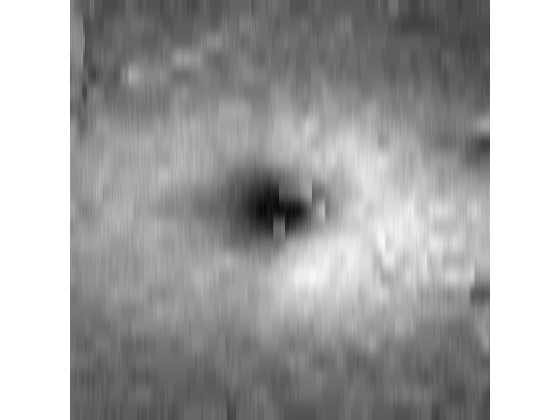} &
 \hspace{-0.5cm}
 \includegraphics[width=2 cm]{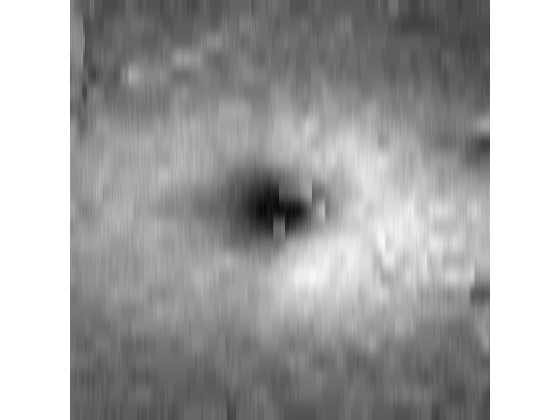} &
 \hspace{-0.5cm}
 \includegraphics[width=2 cm]{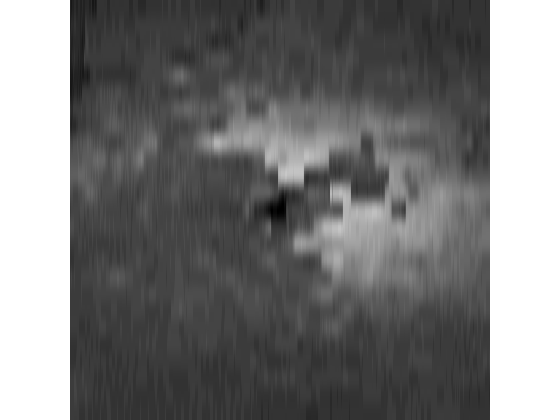} &
 \hspace{-0.5cm}
 \includegraphics[width=2 cm]{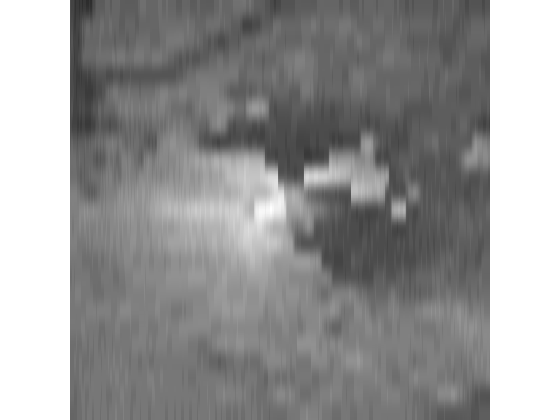} &
 \hspace{-0.5cm}
 \includegraphics[width=2 cm]{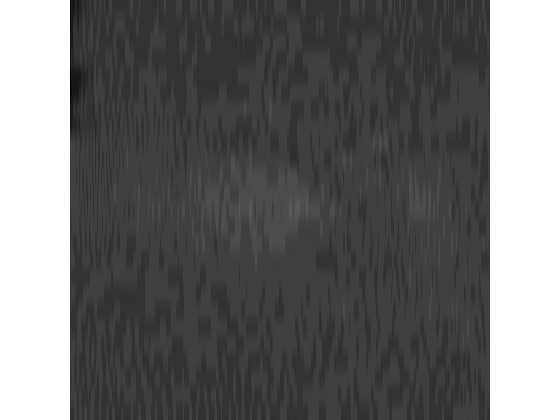} &
 \hspace{-0.5cm}
 \includegraphics[width=2 cm]{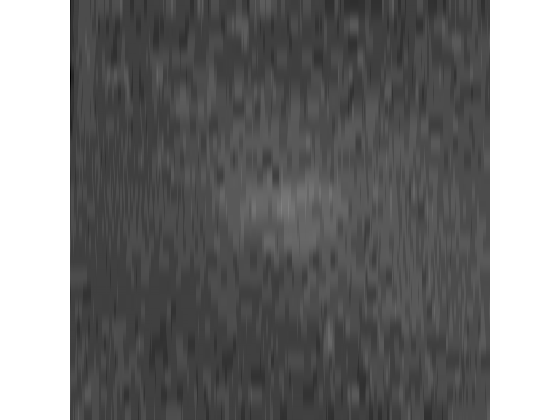} &
 \hspace{-0.5cm}
 \includegraphics[width=2 cm]{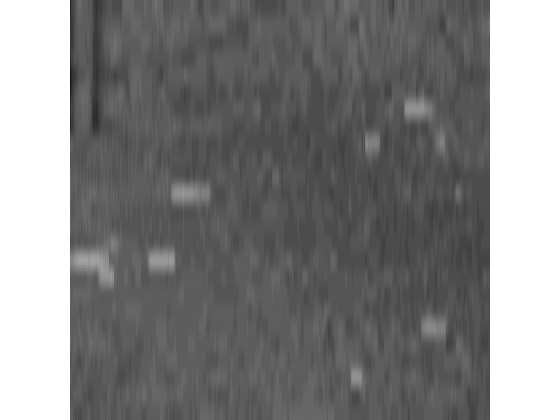} \\
 \includegraphics[width=2 cm]{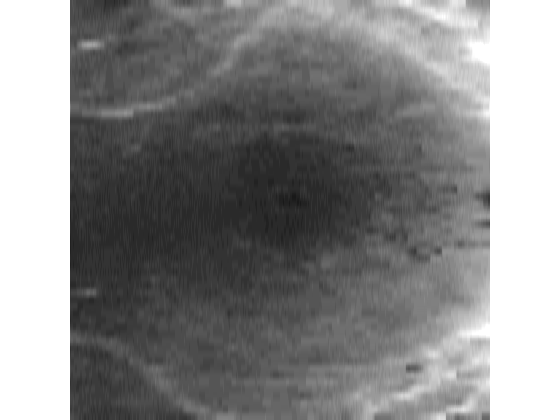} &
 \hspace{-0.5cm}
 \includegraphics[width=2 cm]{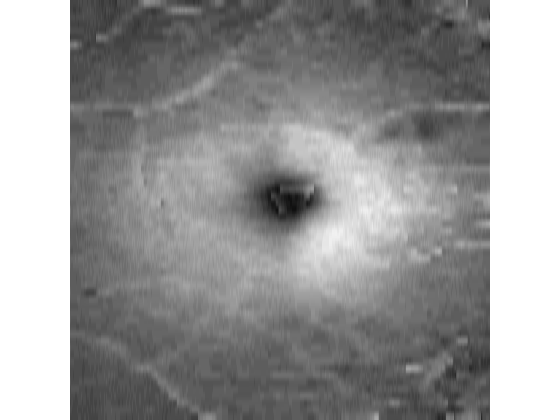} &
 \hspace{-0.5cm}
 \includegraphics[width=2 cm]{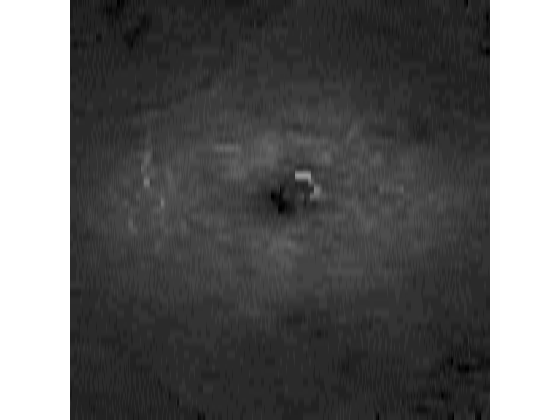} &
 \hspace{-0.5cm}
 \includegraphics[width=2 cm]{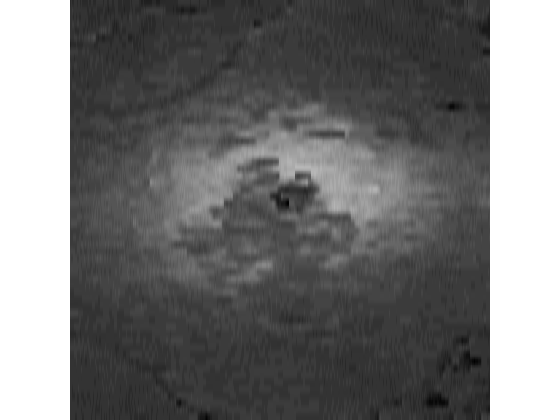} &
 \hspace{-0.5cm}
 \includegraphics[width=2 cm]{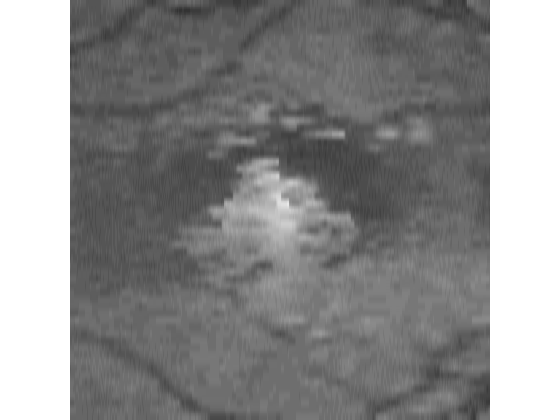} &
 \hspace{-0.5cm}
 \includegraphics[width=2 cm]{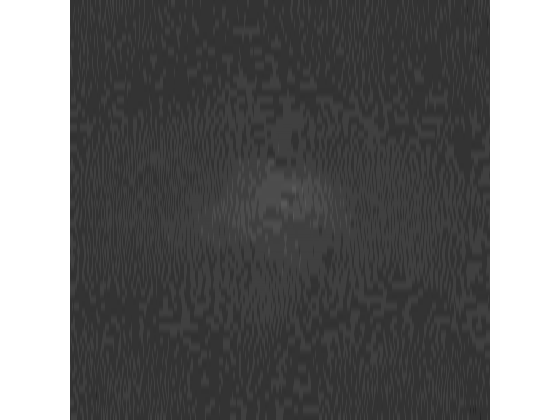} &
 \hspace{-0.5cm}
 \includegraphics[width=2 cm]{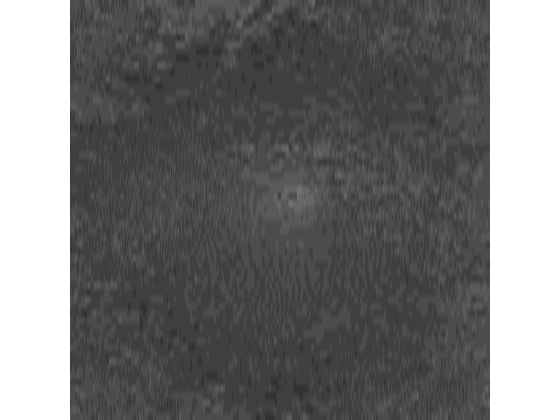} &
 \hspace{-0.5cm}
 \includegraphics[width=2 cm]{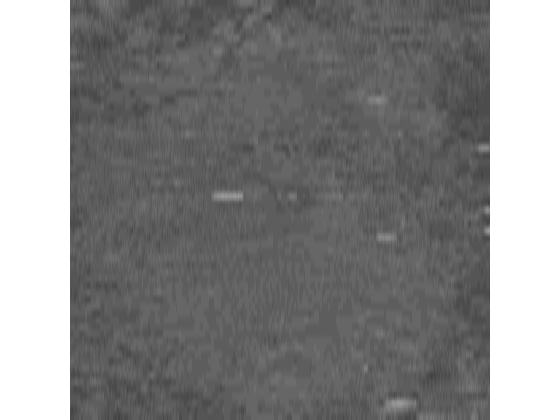} \\
 RNFL & \hspace{-0.5cm} GCL-IPL & \hspace{-0.5cm} INL & \hspace{-0.5cm} OPL & \hspace{-0.5cm} ONL & \hspace{-0.5cm} IS & \hspace{-0.5cm} OS & \hspace{-0.5cm} RPE \\
 \end{tabular}
 \caption{
 MServet FastMac and PPole datasets. Thickness maps of retinal layers obtained from
 the same patient with multiple sclerosis, acquired using the FastMac (top) and PPole (bottom)
 protocols. Details best viewed with zooming.
 }
 \label{fig:ThicknessFastMacPPole}
\end{figure*}

Figure~\ref{fig:ThicknessFastMacPPole} illustrates a qualitative comparison between FastMac and PPole protocols using thickness maps of retinal layers from the same patient with multiple sclerosis.
Both protocols exhibit notable differences in the level of anatomical detail captured, with PPole providing higher spatial resolution.
This increased resolution enables a more precise delineation of retinal layer boundaries and subtle structural variations,
which may be critical for downstream clinical analysis.
These results suggest that PPole acquisitions could offer advantages over FastMac for applications requiring fine-grained structural assessment.

The Supplementary Material provides additional experimental results, including visualizations of topology preservation (Figures S4 - S8), segmentation examples (Figures S9-S13), and layer thickness maps (Figures S14-S18).
The qualitative assessment aligns with the quantitative evaluation, highlighting the robustness of SD-LayerNet-based approaches in preserving anatomically consistent layer structures.
Most differences between method configurations are primarily observed under challenging conditions, where vessels, speckle noise, and shadowing can lead to error propagation across multiple layers.

\section{Discussion}
\label{sec:Discussion}

%
%
%
%
%

\subsection{Spatial normalization in OCT segmentation}

Data augmentation has become the \textit{de facto} standard strategy to address geometric domain shift in medical image
segmentation and other computer vision learning tasks. In this work, we propose an alternative approach based on spatial normalization
to reduce domain shift and improve segmentation performance. Our methodology is inspired by the standard preprocessing
pipelines commonly used in brain MRI analysis, where spatial normalization is a fundamental step in neuroimaging
studies~\cite{Hadj_16}.

Although spatial normalization has rarely been explored within OCT segmentation pipelines, our results indicate that it represents a promising direction
to improve both the robustness of deep-learning segmentation models and the comparability of anatomical retinal regions and quantitative layer measurements.
Despite its widespread use in other imaging domains, normalization in OCT has not been considered as a preprocessing step, and its impact on
layer-wise segmentation performance has not been systematically analyzed.
In this work, we show that appropriate spatial normalization can provide measurable benefits for the simultaneous segmentation of all retinal
layers.


\subsection{SD-LayerNet: best method with no best training configuration}

We compare three segmentation architectures under different training conditions, including models trained on original and
spatially normalized data, without and with data augmentation. Interestingly, models trained on the original dataset with
data augmentation exhibit reasonably good performance when evaluated on spatially normalized test sets. However, the best
results are consistently achieved by models trained directly on normalized data. These findings suggest that spatial
normalization introduces a degree of invariance that facilitates generalization across acquisition conditions.

Among the evaluated methods, SD-LayerNet consistently achieves the best overall performance. However, no single configuration uniformly dominates across all layers
and settings. This reflects the intrinsic heterogeneity of the problem, where the different models behave differently to the varying levels of difficulty in
the images (vessels, artifacts, noise, shadows, etc.), leading to variations in relative performance.




\subsection{Evaluation of segmentation correctness and accuracy}

Our experiments highlight important limitations of commonly used evaluation metrics. In particular, Dice similarity
coefficient (DSC) alone is insufficient to fully assess segmentation quality in this context. In contrast, topology-based
metrics and the analysis of layer thickness maps provide more informative and clinically relevant insights. All evaluation
criteria consistently indicate the superiority of SD-LayerNet, with topology and thickness-based analyses further emphasizing
its robust and stable behavior.
Importantly, our topology-based evaluation metrics are not used in the definition of the loss functions used to train SD-LayerNet.

Our study also highlights an evaluation paradox: scenarios where ground truth annotations are available, and metrics such as DSC can be computed (e.g., JHopkins)
tend to be insufficiently challenging to reveal meaningful differences between methods, whereas in more realistic and complex settings (e.g., MServet)
such annotations are often unavailable.
In this context, the proposed topology-based metrics and thickness-based analysis provide a more sensitive assessment of segmentation quality, enabling the
identification of structural inconsistencies not captured by overlap-based measures.



\subsection{From Heidelberg Measurements to Dense Representations}

From a clinical perspective, Heidelberg devices with OCT acquisition protocols such as FastMac provide measurements
that are highly aggregated in nasal, temporal, inferior, and superior area, which limits their usefulness for anatomically detailed
population studies. PPole protocols offer measurements with improved spatial resolution by providing thickness over
an $8 \times 8$ grid. However, even these representations remain insufficient for a comprehensive and interpretable characterization
of the retinal layer anatomy.

In this context, our results suggest that accurate segmentation methods can recover dense and anatomically meaningful representations
of the thickness of the retinal structures from both protocols, with PPole-derived segmentations offering improved
spatial detail and anatomical consistency. Such detailed information has the potential to be more informative for statistical analyses
and for the development of machine learning and deep learning models aimed at identifying anatomical biomarkers.
This is particularly relevant for the study of neurodegenerative diseases, where subtle structural changes in retinal layers may support diagnosis, prediction,
and allow monitoring of disease progression.

\subsection{Limitations of our study}

A key limitation of the proposed spatial normalization framework is its limited applicability to the peripapillary region. The
normalization strategy adopted in this work is specifically tailored to macular OCT scans, where anatomical consistency across
subjects allows for reliable alignment.
In the macular region, the presence of the fovea provides a well-defined anatomical landmark that can be used to guide spatial
normalization. This enables consistent alignment of B-scans across subjects through translation and rotation, facilitating the
reduction of geometric variability.

In contrast, the peripapillary region lacks an equivalent stable and unique landmark. The optic
disc exhibits considerable inter-subject variability in terms of position, shape, and size, making global alignment significantly more challenging.
Future work will investigate the integration of invertible non-rigid spatial normalization strategies to extend the proposed framework beyond
macular imaging and enable its application to a broader range of OCT acquisition protocols beyond the Heidelberg ecosystem.

\subsection{Our contribution to the state of the art}

An important strength of this study is the evaluation under a cross-domain setting, where models are trained on the JHopkins dataset
under FastMac protocol and tested on the MServet datasets under a different FastMac and PPole protocols. This scenario introduces
a significant spatial domain shift, as the acquisition protocols differ in the retinal regions covered and in their spatial sampling
characteristics. In addition, the MServet dataset reflects real clinical practice and includes subjects with neurodegenerative diseases
and Steinert's disease, which further increases variability due to pathology-related structural alterations.
This is a particularly relevant realistic clinical scenario, where training and deployment data are rarely perfectly matched.

It is also worth noting that related work on MServet datasets has focused on substantially simplified segmentation tasks, such as
isolating the RNFL while grouping the remaining retinal layers into a single class~\cite{Gende_23, Alvarez_24}. Although such approaches may facilitate
extension to more challenging regions such as the peripapillary area, they reduce the intrinsic complexity of the problem by avoiding
the need to accurately delineate multiple thin and closely spaced interfaces.
In contrast, our work addresses full multi-layer retinal segmentation, which requires resolving subtle boundaries and preserving anatomically
consistent topology across all layers. This significantly increases the difficulty of the task with respect to these previous works and places
greater demands on both the segmentation model and the preprocessing pipeline.

\section{Conclusions}
\label{sec:Conclusion}

This work demonstrates that spatial normalization is a key yet underexplored component in OCT segmentation pipelines, with a significant impact on both model robustness and the consistency of downstream retinal analysis. Training and inference in a normalized spatial domain improve generalization across acquisition conditions and facilitate more reliable layer-wise comparisons.
Among the evaluated approaches, SD-LayerNet achieves the best overall performance, particularly when assessed using topology-aware and thickness-based metrics, which better capture clinically relevant structural properties than conventional overlap measures.
Overall, these findings support the integration of spatial normalization as a fundamental element in OCT analysis pipelines and contribute to the development of more robust and clinically meaningful tools for retinal imaging, particularly in the context of neurodegenerative disease research and biomarker discovery.

\section*{Acknowledgments}

This work was partially supported by the national research grants PID2022-138703OB-I00 (Trust-B-EyE project) and
PID2023-148219OB-C22, funded by MCIN/AEI/10.13039/501100011033/FEDER, UE;
RICORS network of inflammatory diseases RD24/0007/0022 form Carlos III Health Institute;
Government of Aragon grant PROY\_B50\_24 and Group Reference T64\_23R (COS2MOS research group);
European HORIZON-MSCA-2024-SE-01 G.A. 10123661;
and European Chistera action CLASiK - PCI2025-167187-2.
The funders had no role in study design, data collection and analysis, decision to publish, or preparation
of the manuscript.

\bibliographystyle{IEEEtran.bst}
\bibliography{abbsmall.bib,OCT.bib}

\end{document}